\documentclass[lettersize,journal]{IEEEtran}
\usepackage{amsmath,amsfonts}
\usepackage{algorithmic}
\usepackage{algorithm}
\usepackage{array}
\usepackage[caption=false,font=normalsize,labelfont=sf,textfont=sf]{subfig}
\usepackage{textcomp}
\usepackage{stfloats}
\usepackage{url}
\usepackage{verbatim}
\usepackage{graphicx}
\usepackage{cite}
\usepackage{amsmath}
\usepackage{lineno}
\usepackage{multirow}
\usepackage[table,xcdraw]{xcolor} 
\usepackage{amsmath,amsfonts}
\usepackage{algorithmic}
\usepackage{booktabs}
\usepackage{graphicx}
\usepackage{textcomp}
\usepackage{hyperref}
\usepackage{graphicx}
\usepackage{setspace}
\usepackage{amssymb}
\usepackage{microtype}
\usepackage{graphicx} 
\usepackage{multirow}
\usepackage{booktabs}

\usepackage{threeparttable}
\usepackage{makecell}
\usepackage{url}
\hypersetup{
hidelinks,
colorlinks=true,
linkcolor=black,
citecolor=black
}

\newcommand{\ie}{\textit{i}.\textit{e}.}
\newcommand{\eg}{\textit{e}.\textit{g}.}

\begin{document}

\title{From Sight to Insight: Unleashing Eye-Tracking in
Weakly Supervised Video Salient Object Detection}

\author{Qi Qin,
        Runmin Cong,~\IEEEmembership{Senior Member,~IEEE,}
         Gen Zhan, Yiting Liao,
and Sam Kwong,~\IEEEmembership{Fellow,~IEEE}
\thanks{This work was supported in part by the Taishan Scholar Project of Shandong Province under Grant tsqn202306079, in part by the the National Natural Science Foundation of China Grant 62471278, and in part by the Research Grants Council of the Hong Kong Special Administrative Region, China Grant STG5/E-103/24-R. (\emph{Corresponding authors: Runmin Cong.})}
\thanks{Qi Qin is with the Institute of Information Science, Beijing Jiaotong University, Beijing 100044, China, and also with the Beijing Key Laboratory of Advanced Information Science and Network Technology, Beijing 100044, China. (e-mail: qiqin96@bjtu.edu.cn)}
\thanks{Runmin Cong is with the School of Control Science and Engineering, Shandong University, Jinan 250061, China, and also with the Key Laboratory of Machine Intelligence and System Control, Ministry of Education, Jinan 250061, China. (e-mail: rmcong@sdu.edu.cn)}
\thanks{Gen Zhan is with ByteDance China, Shenzhen 518000,  China. (e-mail: gen.zhan@bytedance.com)}
\thanks{Yiting Liao is with ByteDance USA, CA 95110, USA. (e-mail: yiting.liao@bytedance.com)}
\thanks{Sam Kwong is with Lingnan University, Hong Kong SAR, China. (e-mail:samkwong@ln.edu.hk)}
}

\markboth{Journal of \LaTeX\ Class Files}%
{Shell \MakeLowercase{\textit{et al.}}: A Sample Article Using IEEEtran.cls for IEEE Journals}


\maketitle

\begin{abstract}
The eye-tracking video saliency prediction (VSP) task and video salient object detection (VSOD) task both focus on the most attractive objects in video and show the result in the form of predictive heatmaps and pixel-level saliency masks, respectively. In practical applications, eye tracker annotations are more readily obtainable and align closely with the authentic visual patterns of human eyes.
Therefore, this paper aims to introduce fixation information to assist the detection of video salient objects under weak supervision.
On the one hand, we ponder how to better explore and utilize the information provided by fixation, and then propose a Position and Semantic Embedding (PSE) module to provide location and semantic guidance during the feature learning process. On the other hand, we achieve spatiotemporal feature modeling under weak supervision from the aspects of feature selection and feature contrast. A Semantics and Locality Query (SLQ) Competitor with semantic and locality constraints is designed to effectively select the most matching and accurate object query for spatiotemporal modeling. In addition, an Intra-Inter Mixed Contrastive (IIMC) model improves the spatiotemporal modeling capabilities under weak supervision by forming an intra-video and inter-video contrastive learning paradigm. Experimental results on five popular VSOD benchmarks indicate that our model outperforms other competitors on various evaluation metrics.
\end{abstract}

\begin{IEEEkeywords}
Video salient object detection, fixation guidance, position and semantic embedding, contrastive learning.
\end{IEEEkeywords}

\section{Introduction}
\IEEEPARstart{V}{ideo} salient object detection (VSOD) task aims to simulate the visual perception mechanism of human eyes and detect the most attractive and moving objects in videos \cite{DBLP:journals/pami/WangLFSLY22}, which has been widely used as a pre-processing step in other fields, such as video compression \cite{DBLP:journals/tip/Itti04}, video semantic segmentation \cite{chen2025replay,fang2025cvpr,DBLP:journals/tcsv/Yang0CWZL24}, video understanding \cite{DBLP:journals/tvcg/WangSYM17}, and video tracking \cite{DBLP:conf/iccv/ZhouP0WZ021}, among others~\cite{cong2025trnet,DBLP:journals/tcyb/CongYLFZHK23,DBLP:journals/tmm/LuoCLIK24,DBLP:journals/tim/CongYJGLWZK22,DBLP:journals/cviu/DongWCSL24}. 
{The closely related task of video saliency prediction (VSP) has also garnered significant attention. The VSP task involves predicting regions of interest in a video, typically represented as heatmaps that highlight areas receiving the most attention, often derived from fixation data. In contrast, the VSOD task focuses on identifying and segmenting these attention-grabbing objects, typically through pixel-level masks. However, there has been limited exploration of joint modeling for the VSP and VSOD tasks. In this work, we aim to combine both tasks, leveraging the VSP network to assist in addressing the VSOD task. Meanwhile, most existing VSOD methods \cite{DBLP:conf/iccv/ZhangLWPYJLLL21} remain fully supervised.}
These methods necessitate pixel-level annotations for every frame of the video, which can be a highly expensive undertaking. To address this problem, some methods have designed the weakly-supervised learning framework \cite{DBLP:conf/cvpr/Zhao0LBLH21,DBLP:conf/mm/GaoXZWGZ22,DBLP:conf/icml/0003C0XH024}, which only needs to annotate a part of the target pixels. Of course, due to incomplete supervision, their performance tends to be lower than fully supervised models, which can be seen as a trade-off between supervisory information and performance.

\begin{figure}[!t]
    \centering
    \includegraphics[scale=0.43]{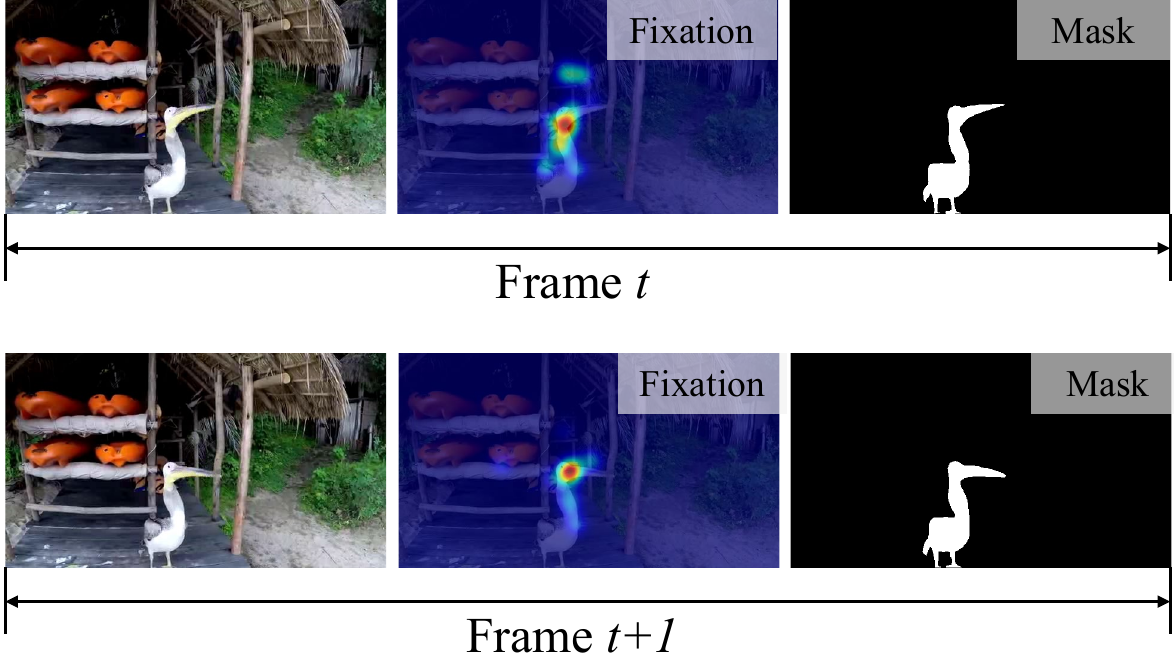}
    \caption{The differences between VSOD task and eye-tracking VSP task. The fixation is the label of the VSP task, and the mask is the label of the VSOD task.}
    \label{fig:fixation_mask}
\end{figure}

For this new attempt, the first thing we need to figure out is, what can fixation information provide for the VSOD task and how should we use it?
Actually, although fixation is closer to the mechanism of human visual attention, there are still some problems that may lead to unreliable results when it is directly applied to the VSOD task.
1) Fixations, whether estimated or captured by eye trackers, are represented as a heatmap-like probability region, only covering a part of the object's area, making it difficult to provide a complete outline of the object for the network. 
2) While fixations can often contain attention-grabbing regions, they often contain pixels that do not belong to the foreground, thus introducing a lot of noise.
Even so, fixation information can provide important location information, which is of great significance for the positioning of salient objects.
To this end, we propose a Position and Semantic Embedding (PSE) module, which utilizes position embedding technology to encode the visual scope represented by fixation information in the network. 
Simultaneously, we introduce a learnable semantic embedding to represent the semantic information of the image. By integrating these two embedding components, the network is equipped with saliency-related position and semantic information.

Another crucial issue is the modeling of spatiotemporal relationships in video sequences. Most previous WVSOD methods use multi-frame relationship modeling or optical flow to learn spatiotemporal features. 
Modeling this relationship is inherently difficult. These methods will perform worse in weakly supervised environments.
Currently, the representative WVSOD methods use supervision in the form of scribble labels \cite{DBLP:conf/cvpr/Zhao0LBLH21} or point labels \cite{DBLP:conf/mm/GaoXZWGZ22}, and both additionally introduce optical flow information to represent spatiotemporal relationships.
In order to enhance the accuracy and efficiency of spatiotemporal modeling, we design the models from the perspective of feature selection and constractive correlation.
On the one hand, the global modeling ability of Deformable DETR as a baseline model is undoubted, but its original object query matching algorithm falls short in effectively localizing salient objects and selecting more comprehensive object features, which may hinder subsequent spatiotemporal feature learning.
{Consequently, incorporating the semantic and locality constraints, we propose a Semantics and Locality Query Competitor (SLQ Competitor) to replace the original object query matching methodology~\cite{carion2020end},
thereby filtering out object queries that better match the targets and providing more accurate features for spatiotemporal modeling.
On the other hand, in order to reduce the computational cost and enhance the ability of feature learning under weakly supervised constraints, we propose an Intra-Inter Mixed Contrastive (IIMC) model to learn spatiotemporal features without introducing the optical flow.}
Specifically, we implement feature constractive learning from two aspects.
1) We construct an intra-video contrast stage by extracting the feature corresponding to the best matching object query between key frame and reference frame, thereby obtaining the precise foreground and background features of each frame. 
2) We design an inter-video contrast stage by using the best matching object query of the key frame and the reference frame as the video-level embedding, thereby preforming inter-video feature contrastive learning between different videos. 
By combining both intra-video and inter-video contrastive learning, our IIMC model enhances the ability of our network to learn discriminative global features, thus improving detection performance of VSOD.

In summary, we propose a Eye-tracking Guided Contrast network (EGCNet) for weakly-supervised video salient object detection and the main contributions are listed as follows: (1) A weakly-supervised video salient object detection called EGCNet is proposed, utilizing the eye-tracking prediction to guide the network learning. Our proposed EGCNet achieves state-of-the-art performance on five public VSOD benchmarks. (2) We propose a PSE module that encodes fixation as a position embedding and combines it with learnable semantic embedding to achieve location-semantic mining for feature learning. (3) We introduce a SLQ Competitor with semantic and locality constraints to effectively select the object query that better matches the target in terms of semantics and localization under weak supervision. (4) We propose an IIMC model including intra-video and inter-video contrastive stages, which improves the spatial-temporal information modeling capabilities under weak supervision and avoids the computational cost caused by optical flow calculations.

\begin{figure*}[!t]
    \centering
    \includegraphics[scale=0.45]{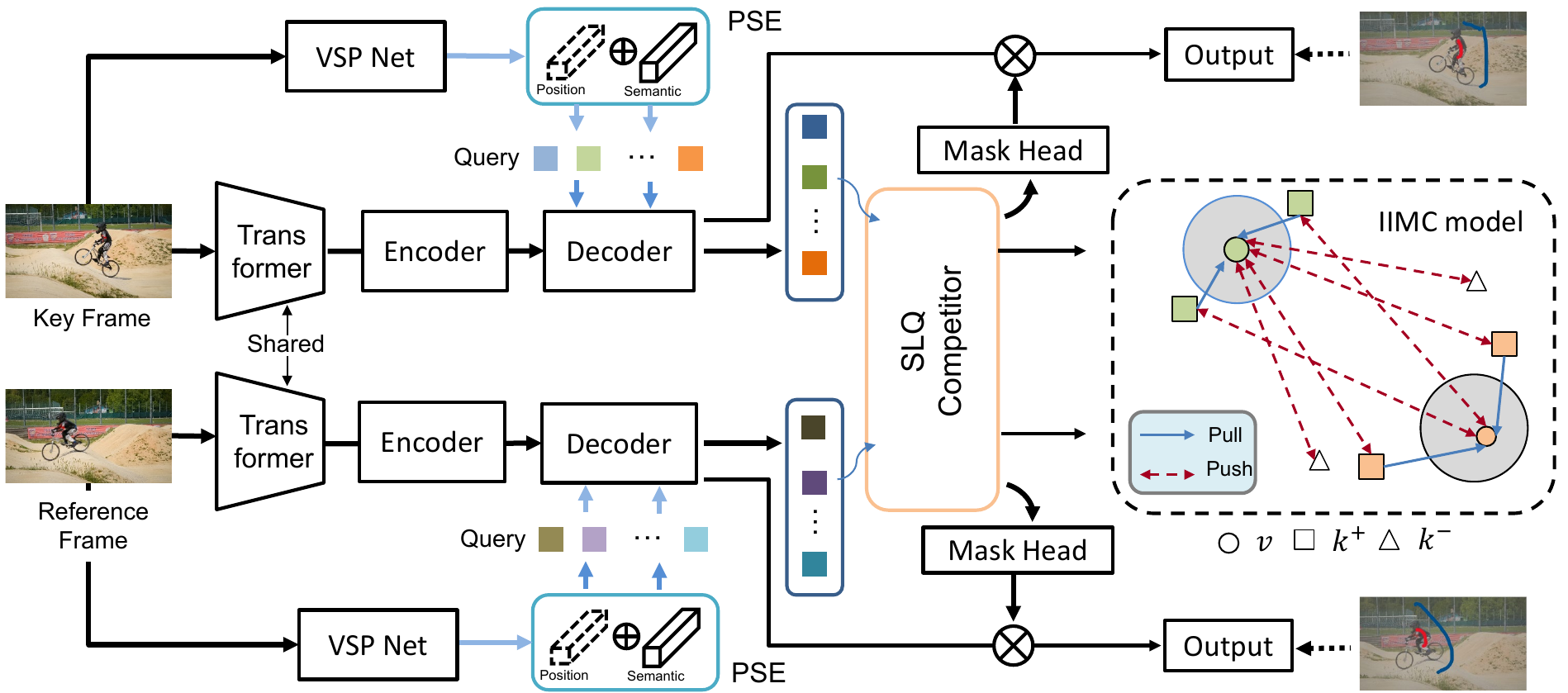}
    \caption{The overall framework of the proposed EGCNet. The PSE module is used to encode the input fixation, providing prior information for network learning. SLQ Competitor is a new object query matching algorithm we designed specifically for the VSOD task. In addition, the IIMC module learns video spatiotemporal relationships by contrasting features from key frame and reference frame.}
    \label{fig:framework}
\end{figure*}

\section{Related work}
\subsection{Weakly Supervised Salient Object Detection for Video/Image }
In recent years, deep learning-based salient object detection methods have achieved excellent performance, but most of them require pixel-level labels for supervision. To reduce the cost of labeling, many weakly supervised or unsupervised methods have been proposed. For single-image salient object detection, \cite{wang2017learning}  first proposed a method that used image-level category labels for supervision and generates rough saliency maps using techniques such as class activation map. \cite{zeng2019multi} combined multiple labels (including image-level category labels and captions) to jointly train a SOD network while also predicting categories and captions. \cite{zhang2020weakly} first used scribble labels for supervision to detect salient object, including using boundary detection tasks to constrain the network and using local cross-entropy to constrain the annotated regions. In addition,  \cite{zhang2018deep} used traditional methods to obtain saliency maps with noise and obtained a clear salient foreground by modeling the noise.  \cite{DBLP:conf/iccv/PiaoWZL21} proposed a novel multi pseudo-label framework that integrates saliency clues from multiple pseudo labels to avoid the inevitable influence of the refinement algorithm adopted for a single label.  \cite{DBLP:conf/aaai/Gao00GZHZ22} used an adaptive flood fill algorithm to expand the point labels and proposed a point-supervised SOD network based on the Transformer. \cite{DBLP:conf/aaai/YanWLZLL22} first proposed an unsupervised domain adaptation SOD method that attempted to adapt from synthetic to real data and constructed a synthetic SOD dataset called UDASOD. \cite{DBLP:journals/tcsv/CongQZJWZK23} proposed a noise label guided weakly supervised SOD framework using hybrid labels (including a small number of pixel-level labels and a large number of noise labels) and proposed a unique training strategy that includes three mechanisms.
It was noteworthy that \cite{DBLP:conf/cvpr/WangSDB18} incorporated fixation into salient object detection, but it only employed fixation as a supervision mechanism, and this was utilized solely in single-image SOD, overlooking the spatiotemporal characteristics inherent between fixations.

For video salient object detection, it is more challenging than single-image salient object detection, and there have also been many deeplearning-based methods proposed.\cite{DBLP:conf/iccv/YanL0LWCL19} proposed a semi-supervised video salient object detection model using pseudo labels and a novel method for generating pixel-level pseudo-labels from sparsely annotated frames.  \cite{DBLP:journals/tcsv/TangZJCHL19} proposed a spatiotemporal cascade neural network architecture for saliency modeling, which utilized limited manually manually labels and a large number of labels generated by existing methods for training. \cite{DBLP:conf/cvpr/Zhao0LBLH21} first used scribble labels in the VSOD task, and re-annotated two commonly used datasets with scribble labels. They also proposed a foreground-background similarity loss and a pseudo-label enhancement method. \cite{DBLP:conf/mm/GaoXZWGZ22} used lower-cost point labels as supervision and proposed a two-stream network with optical flow as input. They re-annotated the training datasets (DAVIS and DAVSOD) using point labels. \cite{zhou2023a2s2} proposed an unsupervised salient object detection model based on texture guidance, utilizing depth, thermal imaging information, and optical flow maps as input. They used object shape information to guide the unsupervised learning process. In contrast to most of the methods that use optical flow as input, we are inspired by a related task (video saliency prediction) and use eye fixation labels as input, which are simpler and more direct to obtain.

\subsection{Video Saliency Prediction}
The purpose of video saliency prediction (VSP) task is to deduce the visual saliency degree of each region in the image, represented in the form of a probability map. In comparison to the VSOD task, the VSP task focuses on the salient regions of the image instead of a specific object. Early VSP models relied on manually crafted spatial and temporal features, but with the emergence of deep learning models, performance has far surpassed traditional methods.
\cite{DBLP:journals/tmm/BakKEE18} constructed a two-stream network structure with optical flow and image as input to learn spatial and temporal features and performed fusion for prediction on the two information sources. \cite{DBLP:conf/cvpr/GorjiC18} proposed a multi-stream ConvLSTM network that included a saliency pathway and three attentional push pathways, which combined the static saliency prediction task for VSP. \cite{DBLP:conf/eccv/JiangXLQW18} utilized an object-to-motion network to extract intra-frame saliency information and applied a ConvLSTM model to model the temporal correlations of inter-frame features. \cite{DBLP:journals/pami/WangSXCLB21} designed a CNN-LSTM network that included an attention module, which, on the one hand, utilized attention mechanism to learn static saliency, and on the other hand, fully learned temporal information using LSTM. They also proposed a high-quality eye-tracking dataset called DHF1K. \cite{DBLP:conf/iccv/MinC19} proposed a VSP network model based on 3D fully convolutional architecture, where the 3D network encoded the spatio-temporal information in a holistic way.
\cite{DBLP:journals/ijcv/BellittoSPRGS21} separately decoded multiple layers of features from the 3D encoder into single-channel saliency maps and integrated them for saliency prediction results.  In general, these methods predict saliency regions and cannot be directly applied to video salient object detection tasks, but VSP can provide important prior knowledge for video salient object detection tasks.

\subsection{Contrastive Learning}
Contrastive learning has achieved great success in image representation learning \cite{DBLP:conf/icml/ChenK0H20,DBLP:conf/cvpr/He0WXG20,DBLP:journals/tmm/CongXCZHZ24,DBLP:journals/tmm/LuoCLIK24,DBLP:conf/cvpr/PangQLCLDY21,DBLP:conf/nips/HanXZ20}. MOCO \cite{DBLP:conf/cvpr/He0WXG20} and  SimCLR \cite{DBLP:conf/icml/ChenK0H20} used contrastive learning to achieve image-level self-supervised training, which provided strong feature representations for downstream tasks. \cite{DBLP:conf/nips/KhoslaTWSTIMLK20} extended contrastive learning to the supervised learning field by treating images with the same label as positive samples and learning better feature representations. \cite{DBLP:conf/cvpr/XieXCHZS22} constructed an unsupervised contrastive segmentation network by assuming that the foreground and background have a large feature-space distance, achieving better results than class activation map (CAM), and providing better activation maps for weakly supervised methods with classification labels as supervision. \cite{DBLP:conf/eccv/WuLJBYB22} learned better spatio-temporal information by using contrastive learning to contrast positive and negative examples from reference frames with target objects from key frames, and applied optimal transport theory to select positive and negative samples to learn more distinctive features. Inspired by these methods, we use contrastive learning to obtain temporal information between videos. By contrasting multiple frames within the same video and across different videos, we learn more distinctive spatio-temporal information.

\section{Method}

The overall framework of our EGCNet is shown in the Fig.\ref{fig:framework}, following an end-to-end network architecture that takes the video $V=\left \{  I_{n} \right \}_{n=1}^{N}$ 
as input to detect salient objects, where $N$ is the length of the video. The Deformable DETR \cite{DBLP:conf/iclr/ZhuSLLWD21} is used as the baseline of our framework. During the training phase, to learn the temporal and motion information of the video, we synchronously input the current key frame ${I_{n}^{key}}$ and a nearby reference frame $I_{n}^{ref}$, and utilize VSP Net \cite{DBLP:conf/iccv/MinC19} to generate corresponding fixations (${Fix_{n}^{key}}$ and ${Fix_{n}^{key}}$) to assist in salient object detection. The VSP Net can be substituted with any existing video saliency prediction model. After feature extraction from the backbone, the features $\left \{ {F}_{3}^{b}, {F}_{4}^{b}, {F}_{5}^{b}\right \}$ extracted by the last three layers of the backbone network are inputted to the Transformer Encoder network for global feature modeling. At the same time, the PSE module is used to introduce the guidance of fixation information. By encoding the position and semantic of the salient regions in the fixation, we obtain the initial values for the object queries of the transformer decoder, accelerating the saliency feature learning. 
After being learned by the decoder, we can obtain the object queries with semantic and positional information of salient objects, where the number of object queries is far greater than the number of salient objects. 
In order to filter out the object queries that best match the salient objects, the object queries are input to the SLQ Competitor for selection, theryby prviding more accurate features for mask prediction via the mask head and contrastive learning via the IIMC moudle. 
Finally, under the weakly scribble supervision, we use the proposed IIMC module to improve the spatial-temporal information modeling capabilities and avoid the computational cost caused by optical flow calculation, including an intra-video contrastive learning and an inter-video contrastive learning on the extracted features from the two frames.

\subsection{Position and Semantic Embedding Module} 

The fixation reflects the areas of interest of human eyes in videos. With the help of fixation, the localization learning of salient object by the network is facilitated. 
As mentioned before, the baseline network we used is an excellent end-to-end paradigm for object detection, but it also has certain shortcomings. In fact, the Deformable DETR relies on the object query in the decoder stage to learn the semantic features and position information of the targets. However, the object query is initialized to an empty state at each training, resulting in longer time to find the objects and slow convergence. Therefore, many methods \cite{DBLP:conf/cvpr/0002Y0Z021} have been proposed to initialize the object query. 
Inspired by these ideas and combined with the physical meaning of fixation, the PSE module does not directly feed fixation and image into the network for feature extraction, but encodes positional and semantic information to provide initialization for object queries.
In this way, on the one hand, the network receives the location prompt from the fixation, and on the other hand, the convergence speed can be accelerated. 

It is worth noting that when encoding fixations, we do not directly use the all pixel positions corresponding to the fixation for position encoding. 
This is because although fixation represents the regions of visual attention, when humans observe a complete video, their gaze does not immediately lock onto salient objects in the first few frames. Instead, it gradually settles on salient objects after several frames.
This can result in the fixation containing a portion of non-salient area in certain frames. 
Therefore, we adopt the representation of a point, \ie, the geometric center of the fixation, as the embedded position of the current salient object. 
After obtaining the position $(x,y)$, we use the spatial position encoding to obtain the position embedding $e_{pos}\in \mathbb{R}^{256}$, which is a simple modification of the original transformer position embedding \cite{DBLP:conf/nips/VaswaniSPUJGKP17} on a 2D image. We achieve expansion on 2D images by performing position embedding separately in the $x$ and $y$ directions, and finally combine the embeddings in the two directions. This process is described as follows:
\begin{align}
     PE(t)=\begin{cases}
  sin(\omega _{k}\cdot t ), \text{ if } i= 2k\\
  cos(\omega _{k}\cdot t ),\text{ if } i=2k+1,
 \end{cases}
\end{align}
\begin{align}
    e_{pos}=Cat(PE(x),PE(y)),
\end{align}
where $\omega _{k}=\frac{1}{10000^{2k/d}}$, $d$ is the encoding dimension, $Cat$ represents concatenation operation.

Then, considering that the features learned by the network should not only contain the position information of the target, but also the semantic features of the object to cope with various complex environments, we add a set of learnable semantic embedding corresponding to the semantic features learned by the object. 
To be specific, we initialize randomly semantic embedding $e_{sem}\in \mathbb{R}^{256}$. Driven by our weak scribble annotations and contrastive objectives, this initially random embedding gradually learns to capture meaningful object semantics. Finally, we combine the semantic embedding and the position embedding into the object query:
\begin{align}
    Q_{ini} = e_{pos} \oplus e_{sem},
\end{align}
where 
$\oplus$ represents element-wise addition operation. The final output $Q_{ini}$ will be used as the object query and input into the decoder of the transformer to learn the position and semantic features of the object.

\subsection{Semantics and Locality Query Competitor}

\begin{figure*}[!t]
    \centering
    \includegraphics[scale=0.55]{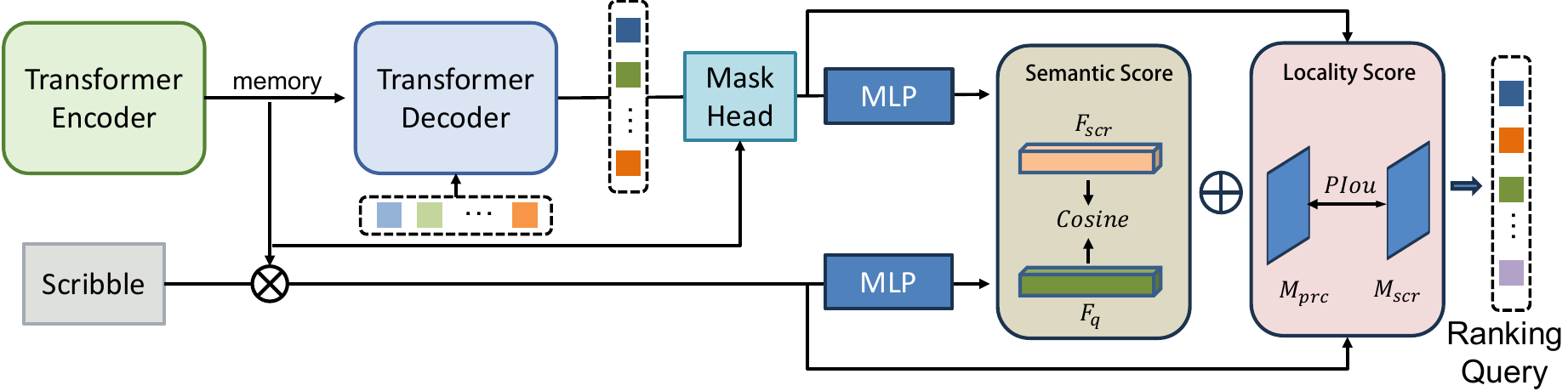}
    \caption{The SLQ Competitor module is employed to assess the object queries and performs evaluations on the generated object query from the decoder based on two key factors: semantic score $Score_{sem}$ and locality score $Score_{loc}$.}
    \label{fig:slq}
    \vspace{-0.3cm}
\end{figure*}

\begin{figure}[!t]
    \centering
    \includegraphics[scale=0.32]{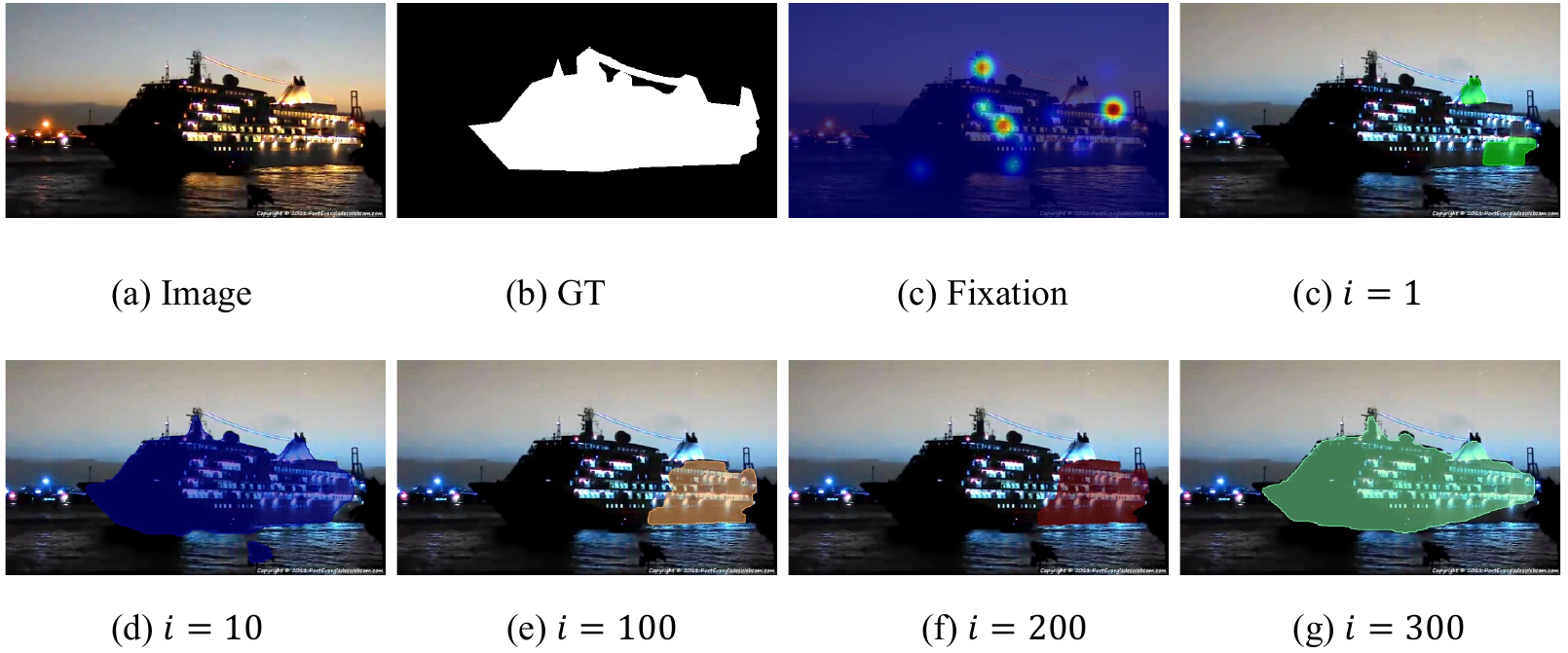}
    \caption{The visualization results of the object queries learned by the decoder are as follows: (a) is the original video frame, (b) is the pixel-level ground truth, (c) is the annotation data collected by the eye tracker. (d)$\sim$(g) are the visualization results of different numbered object queries. $i$ indicates the number of the object query.}
    \label{fig:qeury_vis}
    \vspace{-0.5cm}
\end{figure}

In the baseline network of Deformable DETR, to detect as many categories as possible, it assigns more object queries than the total number of categories. However, each image typically contains a limited number of object categories, making the matching of object queries with their corresponding ground truth a critical challenge. 
Deformable DETR addresses this issue by using the Hungarian algorithm, a bipartite graph matching algorithm. 
This operation is reasonable for object detection task, since images often contain multiple categories of objects.
{However, in the SOD task, the result typically contains objects from a single category (foreground), \ie, the most attractive object, which can be examined as a collective entity.
This implies that we don't need object queries to match multiple ground truths from different categories; Instead, we need to identify the object query that best matches the ground truth. 
 As shown in Fig. \ref{fig:qeury_vis}, the object queries learned by the decoder are not all good. Some of the predicted objects include false positive objects. For example, the tree in Fig. \ref{fig:qeury_vis}(d) is not part of the foreground. In some cases, the detection results are very poor in integrity, as shown in Fig. \ref{fig:qeury_vis}(e).
With this concept in mind, we design a Semantic and Locality Query Competitor (SLQ Competitor) to select the most suitable object query, as shown in Fig. \ref{fig:slq}.}

During the training process of the network, object query not only learns the position information of the salient object, but also learns the semantic information of the object. Therefore, we design two selection criteria to pick out the most matching object query. 
To achieve this goal, we first perform semantic-level filtering. In the Deformable DETR, the encoder outputs a global feature called memory, and object query interacts with it in the decoder to obtain semantic information. Therefore, we first use the provided sparse scribble labels to filter the global information and obtain accurate foreground region features. The process can be described as follows:
\begin{align}
    F_{scr} = NLocal(F_{mem}\otimes M_{scr}+F_{mem}),
\end{align}
where $F_{mem}$ is the global feature (memory), $M_{scr}$ represents scribble annotation, $\otimes$ represents element-wise multiplication operation, and $Nlocal$ represents the Non-local module \cite{DBLP:conf/cvpr/0004GGH18}, which is used to expand the features of the scribble region for later feature comparison. 
Next, we obtain the feature corresponding to all object queries, and compare their similarity with the features of the scribble region. Specifically, we use the dynamic mask head \cite{DBLP:conf/eccv/WuLJBYB22} to obtain the masks corresponding to all object queries, and then use the same operation to filter the memory feature using the mask to obtain the feature $F_{q}$ corresponding to all object queries. The process is shown below:
\begin{align}
    F_{q} = MH(F_{mem}, Query)\otimes F_{mem},
\end{align}
where $MH$ represents Mask Head.  $F_{scr}\in \mathbb{R}^{h \times w}$ and $F_{q}\in \mathbb{R}^{300\times h \times w}$ represent the features filtered by accurate labels and the features corresponding to each object query, respectively, where $h$ and $w$ is the one-fourth of the width and height of the original image, respectively. Using these two sets of features, we can determine the similarity between the semantic information of the object detected by each object query and the accurate foreground features. Thus, we can find the most matching object query based on supervision of semantic information. Overall, we use a MLP layer to map $F_{scr}$ and $F_{q}$ to a unified feature space and adopt cosine similarity to measure the distance between the two features, in order to determine their semantic similarity:
\begin{align}
    Score_{sem} = Cos(MLP(F_{scr}),MLP(F_{q})),
\end{align}
where $Cos$ represents cosine similarity, $MLP$ is the projector function consisting of $1 \times 1$ convolution layer and the activation function ReLU, $Score_{sem}$ denotes sematic score.

Above, we judge the score of each object query from a semantic-level perspective. Next, we evaluate each object query based on its positional information. 
Specifically, we use the map obtained by the mask head for each object query, and compute the Partial-IOU between the foreground/background map and the scribble map as the locality score:
\begin{gather}
     M_{fg} = MH(F_{mem}, Query), M_{bg} = 1-M_{fg},\\
    Score_{loc} = PIou(M_{fg}, M_{scr}^{fg}) + PIou(M_{bg}, M_{scr}^{bg}),
\end{gather}
where 
$M_{fg}$ and $M_{bg}$ refer to the foreground and background mask corresponding to each object query, respectively, $M_{scr}^{fg}$ and $M_{scr}^{bg}$ correspond to the foreground and background masks in the scribble annotation, respectively. $PIou$ is the partial IOU function, which is defined as:
\begin{align}
    PIou(M_{pre}, M_{scr}) = IOU(M_{pre}*M_{scr}, M_{scr}).
\end{align}
Finally, we combine the semantic scores and locality scores, and sort the scores for each object query. The object query with the highest score will be selected as the most matched object query. It can be expressed as follows:
\begin{align}
    Query_{i} = max\left\{ Score_{sem}^{i} + Score_{loc}^{i}\right\}_{i=1}^{N},
\end{align}
where $N=300$ represents the number of object queries.

\subsection{Intra-Inter Mixed Contrastive Model}
\begin{figure*}[!t]
    \centering
    \includegraphics[scale=0.55]{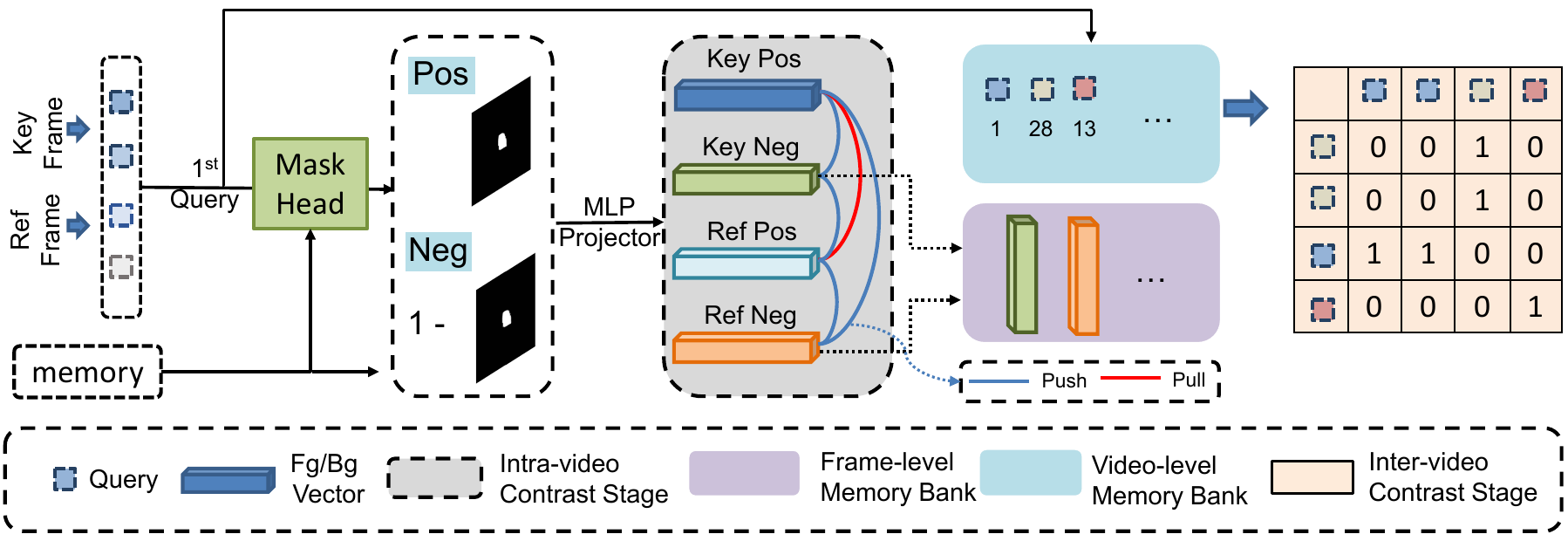}
    \caption{\centering The details of the proposed IIMC model.}
    \label{fig:contrast}
    \vspace{-0.5cm}
\end{figure*}

\vspace{-0.1cm}
As we all know, learning more discriminative inter-frame temporal features can help improve the quality of cross-frame association, which plays a critical role in detecting salient objects.  
As mentioned earlier, establishing this relationship is already difficult, let alone under weak supervision, and optical flow based methods require additional computational costs. Based on these considerations, we introduce contrastive learning to model  spatiotemporal relationships.
Unlike traditional contrastive learning methods that directly perform single-image comparisons after data augmentation, we adopt a dual perspective for contrastive learning in VSOD task, and propose the Intra-Inter Mixed Contrastive Model (IIMC), including the intra-video contrast stage and inter-video contrast stage. 
From a local perspective, we treat different frames within the same video as  individual instances. By comparing the foreground and background within different frames, we aim to improve the target completeness even under weak supervision. From a global perspective, we consider the entire video as a holistic entity, and different frames within the video belong to the same entity. We conduct global comparisons among different videos to reduce the detection of erroneous salient objects under sparse supervision, resulting in more accurate detection results. The IIMC model is illustrated in Fig. \ref{fig:contrast}.

\subsubsection{Intra-video Contrast Stage}

In our framework, object queries are utilized to query the features of the target object in each frame in our pipeline. Hence, the output embeddings can be considered as representing the features of the salient object, including its position and semantic information. 
In this stage, we focus on local perspective for feature learning, enabling the network to achieve good object completeness even under sparse label supervision. Traditional contrastive learning methods often rely on image augmentation to select positive and negative samples, typically based on unsupervised techniques. However, these methods have limitations when it comes to contrasting features, as there may be cases where instances of the same class are incorrectly categorized as negative samples.
Under sparse supervision, we propose a new approach for constructing positive and negative samples, starting from the most matching object query obtained from the SLQ Competitor module. We select positive samples from a local perspective for each frame by using the corresponding features of $Query_{i}$, which represent the foreground features within each frame. The remaining features in each frame are treated as negative samples, representing the background features. We bring positive samples closer to each other while simultaneously pushing positive and negative sample features apart. This approach enables the network to understand the motion variations of salient object among different frames and captures more discriminative spatiotemporal features between foreground and background objects.
Specifically, we employ the foreground activation map $M_{q}$ corresponding to each object query to filter the global features, obtaining the foreground feature representation $V^{fg}$, and use $(1-M_{q})$ as the background activation map to obtain the background feature representation $V^{bg}$, which can be described as follows:
\begin{align}
\abovedisplayskip=12pt
\setlength\abovedisplayskip{0pt}
\setlength\belowdisplayskip{-3pt}
    M_{q} = MH(F_{mem},Query_{i}), 
\end{align}
\begin{align}
     V^{fg} = M_{q} \odot {F_{mem}}^{\top}
\end{align}
\begin{align}
     V^{bg} = (1-M_{q}) \odot {F_{mem}}^{\top}
\end{align}
where $Query_{i}$ is the matched object query by SLQ Competitor, $\odot$ and $\top$ represents the matrix  multiplication  and  transpose, respectively. It is worth noting that $M_{q}$ and $F_{mem}$ are flattened, $M_{q}\in \mathbb{R}^{1\times HW}$ and $F_{mem}\in \mathbb{R}^{C\times HW}$. Meanwhile, $V^{fg} \in \mathbb{R}^{1\times C}$ and $V^{fg} \in \mathbb{R}^{1\times C}$.

For contrastive learning loss, given a key frame $I_{key}$ and a reference frame $I_{ref}$, a foreground representation and a background representation can be generated for each frame, \ie, $(V_{key}^{fg},V_{key}^{bg})$ and $(V_{ref}^{fg},V_{ref}^{bg})$. For $V_{key}^{fg}$, $V_{ref}^{fg}$ belongs to the same salient object and is considered a positive sample, while $V_{key}^{bg}$ and $V_{ref}^{bg}$ are considered negative samples. Likewise, for $V_{ref}^{fg}$, $V_{key}^{bg}$ and $V_{ref}^{bg}$ are negative samples. The intra-video contrastive loss is designed as:
\begin{align}
   \mathcal{L}_{intra} = - \log\frac{\exp(V_{key}^{fg}\cdot V_{ref}^{fg} / \tau )}
{\exp(V_{key}^{fg}\cdot V_{ref}^{fg} / \tau ) + \sum\limits_{a\in A } \sum\limits_{b\in B }\exp(V_{a} \cdot V_{b}/ \tau)},
\label{contras_intra}
\end{align}
Here, the $\cdot$ symbol denotes the inner (dot) product, $\tau \in \mathbb{R}^{+}$ is a scalar temperature parameter, $A = \left \{ V_{key}^{fg}, V_{ref}^{fg} \right \}$, and $B$ includes representations $V_{key}^{bg}$ and $V_{ref}^{bg}$, as well as other background representations from the Frame-level Memory Bank. Under this contrastive learning, if the detection of the current foreground mask $M_{q}$ is incomplete, then $(1-M_{q})$ will inevitably include some foreground objects. This will result in a certain level of similarity between the foreground and background when using contrastive learning. Driven by the constraints of contrastive learning loss, $M_{q}$ will be forced to contain a more complete foreground, while the background mask should contain as few foreground objects as possible.

\subsubsection{Inter-video Contrast Stage}

In the intra-video contrast stage, we conduct comparisons between different frames within the same video. Unlike single-image SOD task where there is no connection between different images, and each image represents a different scene, the difference between frames within a video is minimal in VSOD tasks. Almost all frames in a video are captured in similar scenes. This leads to a lower diversity in video scenes, despite the number of frames in existing datasets being comparable to image datasets (\eg, the DAVSOD dataset contains 23938 frames). Relying solely on local comparisons within the same video may not ensure the robustness of the detection results.
Therefore, we adopt a global perspective by introducing the inter-video contrast stage. In this stage, images from the same video are considered positive samples, while images from different videos serve as negative samples. Additionally, we do not use the same positive and negative sample comparison strategy as the previous stage. Instead, we directly use the $Query_{i}$ from each frame as our positive and negative sample features for comparison from a global viewpoint. This approach allows the intra-video contrast stage to focus more on the overall features of objects, leading to more accurate localization of salient targets. By increasing the richness of the scenes, the generalization ability of the network is further improved.
Specifically, positive samples involve salient objects not only from the key frame and reference frame, but also samples from the same video taken from the Video-level Memory Bank, while negative examples come from different videos in the Video-level Memory Bank. Therefore, inter-video contrastive loss is defined as:
{\begin{align}
    \mathcal{L}_{inter}=\frac{1}{\left | P \right |} \sum\limits _{i^{+}\in P} \mspace{-2mu} - \mspace{-1mu}\log
\frac{\exp(i\cdot i^{+}/\tau)}{\exp(i\cdot i^{+}/\tau)+\sum_{i^{-}\in N}\exp(i\cdot i^{-}/\tau )},
\label{contras_inter}
\end{align}}
where $P$ and $N$ denote query embedding collections of the positive and negative samples, respectively.

{\subsubsection{Memory Bank}
Recent research has indicated that for contrastive learning, a large number of negative samples are crucial as they enable the network to learn more discriminative features. However, the quantity of negative samples is severely limited by the mini-batch size. To address this issue, some methods \cite{DBLP:conf/cvpr/He0WXG20,DBLP:conf/cvpr/WuXYL18} have leveraged large additional memory as a bank to store more negative samples. Inspired by these methods, we have implemented Frame-level and Video-level memory banks for contrastive learning intra-video and inter-video, respectively.

Given a video salient object detection dataset with $N$ frames, our Frame-level Memory Bank is constructed with a size of $N \times D$, where $D$ is the hidden dimension and is set as 256 here. The elements in the Frame-level Memory Bank are D-dimensional feature vectors obtained in the intra-video contrastive learning stage, which include $V_{key}^{bg}$ and $V_{ref}^{bg}$. Constructing the memory bank enables more negative samples to be added to contrastive learning. When computing Eq. (\ref{contras_intra}) for the foreground of the key and reference frame, stored frame-level embeddings with background are viewed as negative samples. 

For the Video-level Memory Bank, its size is $N \times (D+1)$, where $N$ and $D$ still represent the number of frames and hidden dimension, respectively. The elements in the Video-level Memory Bank consist of a Query Embedding and ID $C$. The Query Embedding comes from the most matched object query selected by SLQ competitor, while $C$ represents the video ID of the current frame and is used to construct the relationship matrix for video contrastive learning. When computing Eq. (\ref{contras_inter}), for the current video frame, samples from the same video in the Video-level Memory Bank are considered as positives, while samples from different videos are considered as negatives.

}
\subsection{Inference Strategy}
As our network predicts salient object in an online scheme, where videos are processed frame by frame, predicting a single frame does not fully make use of the temporal information between frames. To address this issue, we develop a memory bank to store historical embedding information. In particular, we initialize an empty memory bank for a test video and perform online prediction for each frame. For each predicted frame, we calculate the matching score $f(i,j)$ between the prediction and the memory bank and find the most matched target $\hat{j}$ for the prediction, following the matching score $f(i,j)$ calculation in \cite{DBLP:conf/eccv/WuLJBYB22}:
\begin{align}
    \hat{j}=\arg \max f(i,j), \forall  j\in \left \{1,2,\dots M  \right \},
\end{align}
where $M$ is number of the embedding in the memory bank. if $f(i,j)>0.1$, we are confident that this prediction is effective. In cases where we cannot find a matching but confident prediction in the memory bank, we embed the prediction into the memory bank.

\section{Experiments}
\subsection{Datasets and Evaluation Metrics}

We conduct experiments on five widely used public VSOD
datasets in order to fully evaluate the effectiveness of our proposed, \ie,  DAVSOD \cite{DBLP:conf/cvpr/FanWCS19}, DAVIS \cite{DBLP:conf/cvpr/PerazziPMGGS16}, SegV2 \cite{DBLP:conf/iccv/LiKHTR13}, FBMS \cite{DBLP:journals/pami/OchsMB14} and ViSal \cite{DBLP:journals/tip/LiXC18}. In this paper, we performed joint training on the DAVIS and DAVSOD training sets. Subsequently, we evaluated our model on the complete SegV2, FBMS, and ViSal datasets, as well as the DAVIS and DAVSOD test sets. We use three widely used evaluation metrics in VSOD task, including  S-measure ($S_{\alpha}$) \cite{fan2017structure}, F-measure ($F_{\beta}$) \cite{DBLP:journals/tcyb/ChenCIK24}, and MAE score~\cite{DBLP:journals/tcsv/Yang0CWZL24}, to compare our proposed method with other state-of-the-art methods.

\subsection{Implementation Details}
\label{details}

\textbf{Network Settings.} We employ Swin-Transformer \cite{DBLP:conf/iccv/LiuL00W0LG21} and ResNet50 as the backbone of our proposed EGCNet and utilize Deformable DETR \cite{DBLP:conf/iclr/ZhuSLLWD21} as the baseline in our paper. For the transformer (Deformabel DETR), we use 6 encoders, 6 decoder layers of width 256, and the number of object queries is set to 300.  Additionally, referring to \cite{DBLP:conf/eccv/WuLJBYB22}, we also incorporate a dynamic mask head in our approach. 

\textbf{Training Settings.} In our approach, {we begin by pre-training on the S-DUTS dataset, utilizing scribble annotations, without incorporating DETR pre-trained weights. The images are randomly cropped, ensuring that the shortest side measures no less than 320 pixels and no more than 640 pixels.}
During fine-tuning by video datasets, we utilize TASED-Net \cite{DBLP:conf/iccv/MinC19} to generate fixations with video inputs. During training, we designate the current frame as the key frame and randomly select a frame within the 10 surrounding frames as the reference frame. The fixations of both frames are then used as inputs for the network during training. We rely on Scribble annotations, as described in \cite{DBLP:conf/cvpr/Zhao0LBLH21}, for supervised information. At the same time, we use the same scale augmentation with S-DUTS to resize the inputs to a maximum of 576 and a minimum of 320. We use the AdamW \cite{DBLP:conf/iclr/LoshchilovH19} optimizer with base learning rate of $1 \times 10^{-4}$,
and weight decay of $10^{-4}$. We train our network for 12000 iterations for video datasets and reduce the learning rate by a factor of 10 at the 8,000 iteration. {The model based on Transformer is trained on 8 V100 GPUs of 32G RAM, with 2 pairs of frames per GPU. Furthermore, our ResNet-based variant is trained on a single 32GB V100 GPU.}

\subsection{Comparison With the State-of-The-Arts}
To prove the effectiveness of our proposed EGCNet, we compair with 20 state-of-the-art models, including  2 single image fully supervised SOD methods (\ie, EGNet \cite{DBLP:conf/iccv/ZhaoLFCYC19} and PoolNet \cite{DBLP:conf/cvpr/LiuHCFJ19}), 13 fully supervised VSOD methods (\ie, MBNM \cite{DBLP:conf/eccv/LiSVLK18}, PDB \cite{DBLP:conf/eccv/SongWZSL18}, FGRN \cite{DBLP:conf/cvpr/Li0WWL18}, MGA \cite{DBLP:conf/iccv/LiCLY19}, RCRNet \cite{DBLP:conf/iccv/YanL0LWCL19}, SSAV \cite{DBLP:conf/cvpr/FanWCS19}, PCSA \cite{DBLP:conf/aaai/GuWW0CL20}, TENet \cite{DBLP:conf/eccv/RenH0HH20}, DCFNet \cite{DBLP:conf/iccv/ZhangLWPYJLLL21}, MQP \cite{DBLP:journals/tcsv/ChenSPWF22}, FSNet (\cite{DBLP:conf/iccv/JiFWFS021}), PSNet \cite{DBLP:journals/tetci/CongSLYZK23} and UFO \cite{DBLP:journals/tmm/SuDSLSW24}) and 5 weakly supervised/unsupervised VSOD methods (\ie, GF \cite{DBLP:journals/tip/WangSS15}, WSSA \cite{DBLP:conf/cvpr/ZhangYLSLD20}, WVSOD
 \cite{DBLP:conf/cvpr/Zhao0LBLH21}, WVSODP \cite{DBLP:conf/mm/GaoXZWGZ22} and A2Sv2 \cite{zhou2023a2s2}). For a fair comparison, all the results are provided
directly by the authors or generated by the source codes
under the default parameter settings in the corresponding
models.
\begin{figure*}[!h]
    \centering
    \includegraphics[scale=0.51]{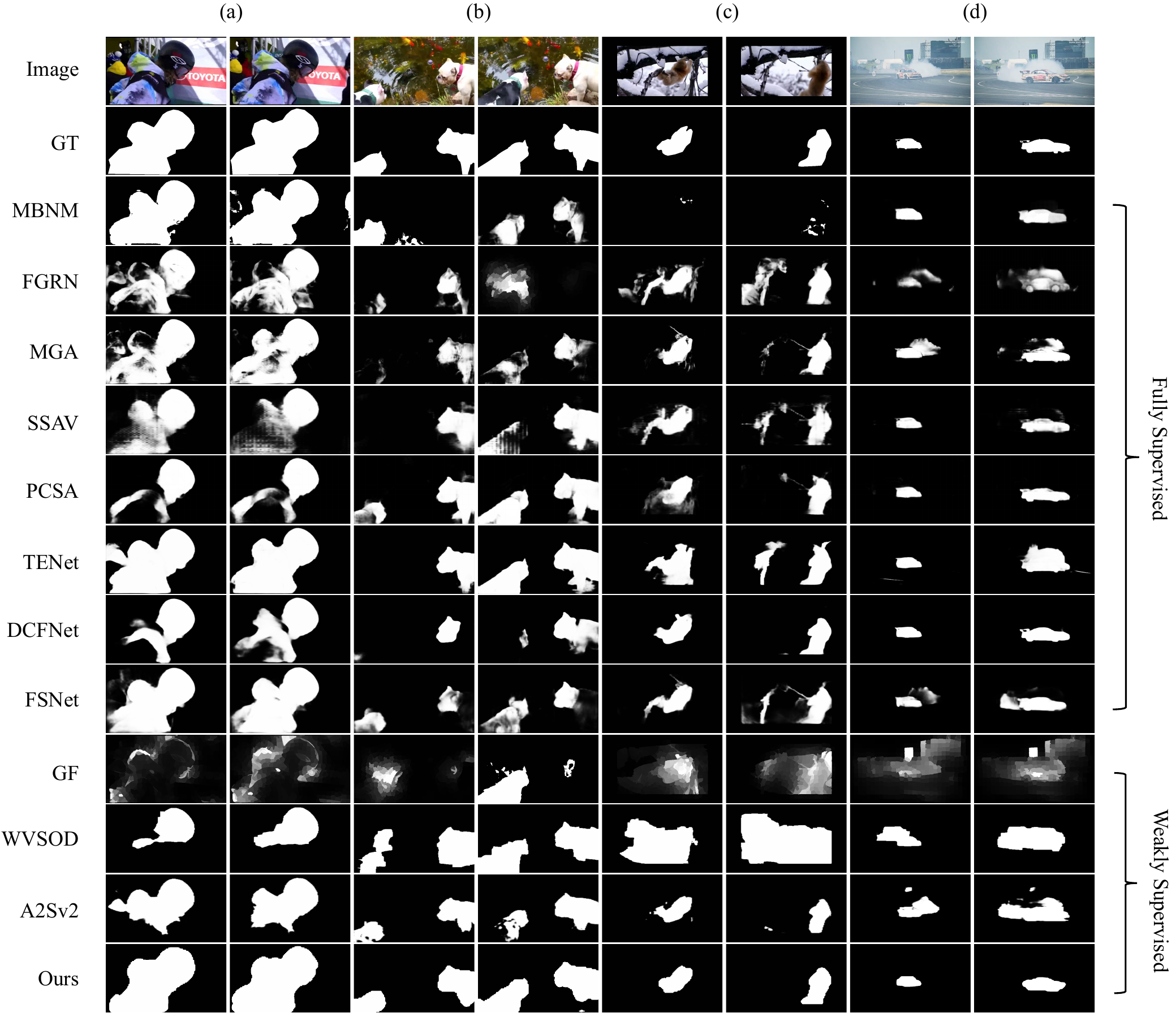}
    \caption{Qualitative comparison with other methods. As can be seen from rows 3 to 10 (fully supervised methods) and 11 to 14 (weakly supervised methods), our approach outperforms other weakly supervised methods in terms of completeness, localization accuracy, and complex environment detection.}
    \label{fig:compare_fig}
\end{figure*}
\subsubsection{Qualitative Evaluation}
Fig. \ref{fig:compare_fig} presents some challenging examples, including visualizations of various scenes and resolutions. These results not only demonstrate the superior performance of our approach in terms of localization accuracy and completeness of salient objects compared to other weakly supervised methods, and even some fully supervised methods, but also highlight its detection capability in complex scenes, such as those with low contrast, complex backgrounds, and severe occlusions.

\textbf{1) Advantages in Integrity Detection.} In comparison to other methods for video salient object detection, our EGCNet exhibits strong performance in detecting the completeness of salient objects, even surpassing some fully supervised methods. For example, in Fig. \ref{fig:compare_fig}(a), our approach outperforms other weakly supervised and some fully supervised methods by detecting the person more holistically. Notably, our method effectively identifies the complete attire of the person, including the hat, while some methods (\ie, PCSA, DCFNet and WVSOD) fail to do so. 
In Fig. \ref{fig:compare_fig}(b), the alternating black and white coloration of the dog on the left presents a challenge for some weakly supervised methods (\ie, WVSOD and A2Sv2), resulting in the failure to detect the full body of the dog. Additionally, salient objects that deviate from the image center can cause some fully supervised methods (\ie, MBNM, DCFNet and TENet) to lose track of them entirely in the detection results.

\begin{table*}[!t]
    \caption{Quantitative results on the VSOD dateset, including  DAVSOD \cite{DBLP:conf/cvpr/FanWCS19}, DAVIS  \cite{DBLP:conf/cvpr/PerazziPMGGS16}, SegV2 \cite{DBLP:conf/iccv/LiKHTR13}, FBMS \cite{DBLP:journals/pami/OchsMB14}, and ViSal \cite{DBLP:journals/tip/LiXC18}. The best result is marked in \textcolor{red}{\textbf{RED}}, and the second best result is marked in \textcolor{blue}{BLUE}. 
 \boldmath $\dag$ indicates that Transformer is used as backbone, and all other methods are based on ResNet50. "OF" stands for whether to use optical flow as input.
 'Un', 'S' and 'P' respectively denote the methodologies of unsupervision, scribble-based supervision and point-based supervision.}
 
	\begin{center}
				 \renewcommand\arraystretch{1.5}
		\resizebox{\textwidth}{!}{
			\begin{tabular}{c|c|c|c|ccc|ccc|ccc|ccc|ccc}
			\toprule
\multicolumn{1}{c|}{\multirow{2}{*}{Method}} 
& \multicolumn{1}{c|}{\multirow{2}{*}{Pub'Year}} 
& \multicolumn{1}{c|}{\multirow{2}{*}{Sup}} 
& \multicolumn{1}{c|}{\multirow{2}{*}{OF}} 
& \multicolumn{3}{c|}{\textbf{DAVIS}}   
& \multicolumn{3}{c|}{\textbf{DAVSOD}}   
& \multicolumn{3}{c|}{\textbf{SegV2}} 
& \multicolumn{3}{c|}{\textbf{FBMS}} 
& \multicolumn{3}{c}{\textbf{ViSal}} 
\\ \cline{5-19} 
\multicolumn{1}{c|}{} 
&\multicolumn{1}{c|}{}   
&\multicolumn{1}{c|}{} 
&\multicolumn{1}{c|}{} 
&\multicolumn{1}{c}{$S_{\alpha}\uparrow$} &\multicolumn{1}{c}{$F_{\beta}\uparrow$}  &\multicolumn{1}{c|}{${MAE}\downarrow$} 
&\multicolumn{1}{c}{$S_{\alpha}\uparrow$} &\multicolumn{1}{c}{$F_{\beta}\uparrow$}  &\multicolumn{1}{c|}{${MAE}\downarrow$} 
&\multicolumn{1}{c}{$S_{\alpha}\uparrow$} &\multicolumn{1}{c}{$F_{\beta}\uparrow$}  &\multicolumn{1}{c|}{${MAE}\downarrow$} 
&\multicolumn{1}{c}{$S_{\alpha}\uparrow$} &\multicolumn{1}{c}{$F_{\beta}\uparrow$}  &\multicolumn{1}{c|}{${MAE}\downarrow$} 
&\multicolumn{1}{c}{$S_{\alpha}\uparrow$} &\multicolumn{1}{c}{$F_{\beta}\uparrow$}  &\multicolumn{1}{c}{${MAE}\downarrow$} 
\\ \toprule

                    EGNet$\ast$ \cite{DBLP:conf/iccv/ZhaoLFCYC19} &ICCV'19 &Ful
                    &
                    & 0.829 & 0.768 & 0.057
                    & 0.719 & 0.604 & 0.101
                    & 0.845 & 0.774 & 0.024
                    & 0.878 & 0.848 & 0.044
                    & 0.946 & 0.941 & 0.015 \\

                    PoolNet$\ast$ \cite{DBLP:conf/cvpr/LiuHCFJ19} & CVPR'19 &Ful
                    &
                    & 0.854 & 0.815 & 0.038
                    & 0.702 & 0.592 & 0.089
                    & 0.782 & 0.704 & 0.025
                    & 0.839 & 0.830 & 0.060
                    & 0.902 & 0.891 & 0.025\\
\hline
				
                    MBNM \cite{DBLP:conf/eccv/LiSVLK18} & ECCV'18 &Ful & \checkmark
                    & 0.887 & 0.861 & 0.031
                    & 0.637 & 0.520 & 0.159
                    & 0.809 & 0.716 & 0.026
                    & 0.857 & 0.816 & 0.047
                    & 0.898 & 0.883 & 0.020\\

                    PDB \cite{DBLP:conf/eccv/SongWZSL18} & ECCV'18 &Ful &
                    & 0.882 & 0.855 & 0.028
                    & 0.698 & 0.572 & 0.116
                    & 0.864 & 0.808 & 0.024
                    & 0.851 & 0.821 & 0.064
                    & 0.907 & 0.888 & 0.032\\

                    FGRN \cite{DBLP:conf/cvpr/Li0WWL18} &CVPR'18 &Ful & \checkmark
                    & 0.838 & 0.783 & 0.043
                    & 0.693 & 0.573 & 0.098
                    & -     & -     & -
                    & 0.809 & 0.767 & 0.088
                    & 0.861 & 0.848 & 0.045\\

                    MGA \cite{DBLP:conf/iccv/LiCLY19} &ICCV'19 &Ful & \checkmark
                    & 0.910 & 0.892 & 0.023
                    & 0.741 & 0.643 & 0.083
                    & 0.880 & 0.829 & 0.027
                    & \textcolor{blue}{0.908} & \textcolor{blue}{0.903} & \textcolor{blue}{0.027}
                    & 0.940 & 0.936 & 0.017\\

                    RCRNet \cite{DBLP:conf/iccv/YanL0LWCL19} &ICCV'19 &Ful & \checkmark
                    & 0.886 & 0.848 & 0.027
                    & 0.741 & 0.654 & 0.087
                    & 0.843 & 0.782 & 0.035
                    & 0.872 & 0.859 & 0.053
                    & 0.922 & 0.907 & 0.026\\

                    SSAV \cite{DBLP:conf/cvpr/FanWCS19} &CVPR'19  &Ful &
                    & 0.892 & 0.860 & 0.028
                    & 0.755 & 0.659 & 0.084
                    & 0.849 & 0.797 & 0.023
                    & 0.879 & 0.865 & 0.040
                    & 0.942 & 0.938 & 0.021\\

                    PCSA \cite{DBLP:conf/aaai/GuWW0CL20} &AAAI'20 &Ful &
                    & 0.902 & 0.880 & 0.022
                    & 0.741 & 0.656 & 0.086
                    & 0.866 & 0.811 & 0.024
                    & 0.868 & 0.837 & 0.040
                    & 0.946 & 0.941 & 0.017\\

                    TENet \cite{DBLP:conf/eccv/RenH0HH20} &ECCV'20 &Ful & \checkmark
                    & 0.905 & 0.881 & 0.017
                    & \textcolor{red}{\textbf{0.779}} & \textcolor{blue}{0.697} & \textcolor{red}{\textbf{0.070}}
                    & 0.868 & 0.810 & 0.025
                    & \textcolor{red}{\textbf{0.916}} & \textcolor{red}{\textbf{0.915}} & \textcolor{red}{\textbf{0.024}}
                    & 0.949 & 0.949 & \textcolor{blue}{0.012}\\

                    DCFNet \cite{DBLP:conf/iccv/ZhangLWPYJLLL21} &ICCV'21 &Ful &
                    & 0.914 & 0.900 & \textcolor{blue}{0.016}
                    & 0.755 & 0.660 & 0.074
                    & \textcolor{red}{\textbf{0.893}} & 0.837 & \textcolor{red}{\textbf{0.014}}
                    & -     & -     & -
                    & 0.952 & \textcolor{blue}{0.953} & \textcolor{red}{\textbf{0.010}}\\

                    MQP \cite{DBLP:journals/tcsv/ChenSPWF22} &TCSVT'21 &Ful & \checkmark
                    & 0.916 & 0.904 & 0.018
                    & 0.770 & \textcolor{red}{\textbf{0.703}} & 0.075
                    & 0.882 & \textcolor{blue}{0.841} & 0.018
                    & -     & -     & -
                    & 0.942 & 0.939 & 0.016 \\

                    FSNet \cite{DBLP:conf/iccv/JiFWFS021}  & ICCV'21 &Ful &\checkmark
                    & \textcolor{red}{\textbf{0.920}} & \textcolor{red}{\textbf{0.907}} & \textcolor{blue}{0.016}
                    & \textcolor{blue}{0.773} & 0.685 & \textcolor{blue}{0.072}
                    & 0.870 & 0.805 & 0.024
                    & -     & -     & -
                    & -     & -     & -    \\

                    PSNet \cite{DBLP:journals/tetci/CongSLYZK23} &TETCI'22 &Ful  & \checkmark
                    & \textcolor{blue}{0.919} & \textcolor{red}{\textbf{0.907}} & \textcolor{blue}{0.016}
                    & 0.765 & 0.678 & 0.074
                    & \textcolor{blue}{0.889} & \textcolor{red}{\textbf{0.852}} & \textcolor{blue}{0.016}
                    & -     & -     & -
                    & \textcolor{blue}{0.954} & \textcolor{red}{\textbf{0.955}} & \textcolor{blue}{0.012}\\

                    UFO \cite{DBLP:journals/tmm/SuDSLSW24}& TMM'24 &Ful &\checkmark
                    & 0.918 & \textcolor{blue}{0.906} & \textcolor{red}{\textbf{0.015}}
                    & -     & -     & -
                    & -     & -     & -
                    & 0.891 & 0.888 & 0.031
                    & \textcolor{red}{\textbf{0.959}} & 0.951 & 0.013 \\

                    \hline
                    
                     GF \cite{DBLP:journals/tip/WangSS15} &TIP'15 &Un & \checkmark
                    & 0.688 & 0.569 & 0.100
                    & 0.553 & 0.334 & 0.167
                    & 0.699 & 0.592 & 0.091
                    & 0.651 & 0.571 & 0.160
                    & 0.757 & 0.683 & 0.107\\

                    WSSA \cite{DBLP:conf/cvpr/ZhangYLSLD20} &CVPR'20 &S & 
                    & 0795 & 0.734 & 0.044
                    & 0.672 & 0.556 & 0.101
                    & 0.733 & 0.664 & 0.039
                    & 0.747 & 0.727 & 0.083
                    & 0.853 & 0.831 & 0.038\\

                    WVSODS \cite{DBLP:conf/cvpr/Zhao0LBLH21} &CVPR'21 &S & \checkmark
                    & \textcolor{red}{\textbf{0.828}} & 0.779 & 0.037
                    & 0.705 & 0.605 & 0.103
                    & 0.804 & 0.738 & 0.033
                    & 0.778 & 0.786 & 0.072
                    & 0.857 & 0.831 & 0.041\\

                    WVSODP \boldmath $\dag$ \cite{DBLP:conf/mm/GaoXZWGZ22} &MM'22 &P & \checkmark
                    & 0.808 & 0.754 & 0.040
                    &\textcolor{blue}{0.718} & \textcolor{blue}{0.622} & 0.094
                    & \textcolor{blue}{0.834} & 0.753 & 0.030
                    & \textcolor{blue}{0.812} & 0.794 & 0.055
                    & \textcolor{blue}{0.878} & \textcolor{blue}{0.858} & \textcolor{blue}{0.028}\\

                    A2Sv2 \cite{zhou2023a2s2} &CVPR'23 &Un & \checkmark
                    & 0.781	& 0.752	& 0.043
                    & 0.633	& 0.571	& 0.105
                    & 0.751	& 0.745	& 0.033
                    & 0.698	& 0.736	& 0.098
                    & -     & -     & -   \\
                    EGCNet &- &S & 
                    &0.816	&\textcolor{blue}{0.804}	&\textcolor{blue}{0.036}	
                    &0.713	&0.612	&\textcolor{blue}{0.093}	
                    &0.823	&\textcolor{blue}{0.785}	&\textcolor{blue}{0.028}	
                    &0.805	&\textcolor{blue}{0.793}	&\textcolor{blue}{0.049}	
                    &0.865	&0.842	&0.029\\
                    EGCNet \boldmath $\dag$ &- &S &
                    & \textcolor{blue}{0.826}	& \textcolor{red}{\textbf{0.821}}	& \textcolor{red}{\textbf{0.032}}
                    & \textcolor{red}{\textbf{0.737}}	& \textcolor{red}{\textbf{0.661}}	& \textcolor{red}{\textbf{0.076}}	
                    & \textcolor{red}{\textbf{0.838}}	& \textcolor{red}{\textbf{0.806}}	& \textcolor{red}{\textbf{0.019	}}
                    & \textcolor{red}{\textbf{0.824}}	& \textcolor{red}{\textbf{0.803}}	& \textcolor{red}{\textbf{0.046}}	
                    & \textcolor{red}{\textbf{0.880}}	& \textcolor{red}{\textbf{0.875}}	& \textcolor{red}{\textbf{0.026}}\\
				\bottomrule
			\end{tabular}}
	\end{center}
	\label{compare_table}
\end{table*}

\textbf{2) Advantages in Background Suppression.}
Although single-image salient object detection has achieved high accuracy, when applied to video salient object detection, variation across consecutive frames can lead to incorrect results. Detecting wrong salient objects can mislead subsequent frames in online models, resulting in cascading detection errors. Our model achieves higher accuracy than other methods by suppressing background noise (\ie, non-salient objects). For example, in Fig. \ref{fig:compare_fig}(c), some methods (\ie, FGRN, MGA, SSAV, FSNet and A2Sv2) detect a spurious branch due to its proximity to the monkey, while others (\ie, MBNM and WVSOD) fail to detect the salient object due to heavy snow. 

\textbf{3)  Advantages in Complex and Challenging Scenarios.}
Compared to other weakly-supervised VSOD methods, our EGCNet is capable of handling various complex scenarios, including situations with low contrast, complex backgrounds, and severe occlusions. For example, in Fig. \ref{fig:compare_fig}(d), smoke and vehicles as well as the surrounding environment have a similar color, leading to weakly supervised (\ie, GF, WVSOD and A2Sv2) or even fully supervised methods (\ie, FGRN, MGA, TENet and FSNet) mistakenly classifying smoke as salient objects. Additionally, some smoke is closely attached to object edges, further complicating the classification task. Furthermore, these methods fail to learn how to exclude smoke across multiple frames. In contrast, our method effectively addresses background interference issues, particularly with regards to smoke.

{\subsubsection{Quantitative Evaluation}
To provide a more intuitive comparison, we present Table \ref{compare_table} which shows the quantitative results of our proposed EGCNet. The best and second-best performers are marked in red and blue for a better visual comparison. We conduct comprehensive performance comparisons against single-image salient object detection methods as well as fully and weakly supervised video salient object detection methods on five widely-used datasets, including S-measure ($S_{\alpha}$), F-measure ($F_{\beta}$), and MAE score. Additionally, we indicate whether optical flow maps were used as inputs, denoted as "OF." 

As illustrated in Table \ref{compare_table}, under a relatively fair condition, our method achieves state-of-the-art results on the majority of evaluation metrics, with only one metric showing slight limitations. For example, on the DAVIS dataset, our method shows a 5.4\% improvement in F-measure, increasing from 0.779 to 0.821, and a decrease in MAE from 0.037 to 0.032 with a 13.5\% improvement to the second-best weakly supervised video salient object detection method. On the DAVSOD dataset, our approach excels in all evaluation metrics, including a 2.65\% increase in S-score from 0.718 to 0.737, a 6.27\% increase in F-score from 0.622 to 0.661, and a decrease in MAE from 0.094 to 0.076 with a 19.1\% improvement, even outperforming some fully supervised methods (such as PCSA and SSAV).

In addition, we provide a ResNet-based variant of our model. Compared to other ResNet-based weakly supervised video salient object detection methods (WSSA, WVSODS, WVSODP and A2Sv2t), our approach consistently outperforms them across nearly all evaluation metrics. Specifically, on the DAVIS and DAVSOD datasets, our model exceeds the performance of the previous state-of-the-art ResNet-based method WVSODS by 3.2\% and 1.2\% in F-measure, respectively. Notably, our method also surpasses certain Transformer-based approaches, such as WVSODP, on specific metrics. For example, on DAVSOD and SegV2, our model achieves improvements of 10.8\% and 15.2\% in MAE, respectively.
}

\subsection{Ablation Study}
In this section, we conduct experiments to evaluate the effectiveness of three modules, including the PSE, SLQ Competitor, and IIMC modules. We perform the experiments on the DAVIS and DAVSOD datasets, respectively.

\subsubsection{Effectiveness of PSE module}
To verify the effectiveness of the proposed PSE module, we conduct several experiments in Table \ref{ab-psem} with the following experimental settings:

\begin{table}[!t]
	\scriptsize
	\caption{Quantitative ablation evaluation of PSE module on the VSOD datasets DAVIS and DAVSOD. Black bold fonts indicate the best performance.}
	\begin{center}
				 \renewcommand\arraystretch{1.5}
		\setlength{\tabcolsep}{1.8mm}{
		\vspace{0.2cm}
			\begin{tabular}{c|ccc|ccc}
			\toprule
\multicolumn{1}{c|}{\multirow{2}{*}{Method}} & \multicolumn{3}{c|}{\textbf{DAVIS}}  & \multicolumn{3}{c}{\textbf{DAVSOD}}                                                                                                                                                                 \\ \cline{2-7} 

\multicolumn{1}{c|}{}
&\multicolumn{1}{c}{$S_{\alpha}\uparrow$} 
&\multicolumn{1}{c}{$F_{\beta}\uparrow$} & \multicolumn{1}{c}{${MAE}\downarrow$} 
& \multicolumn{1}{c}{$S_{\alpha}\uparrow$}  
&\multicolumn{1}{c}{$F_{\beta}\uparrow$}
& \multicolumn{1}{c}{$MAE\downarrow$} \\ \toprule
w/o Sem
& 0.815	 & 0.804	& 0.035
& 0.729  & 0.653    & 0.080      \\
w/o Pos 
& 0.794  & 0.783	& 0.048
& 0.713  & 0.632    & 0.092      \\
tra-fix
& 0.812 & 0.804	    & 0.037
& 0.721 & 0.642     & 0.083     \\
FULL 
&\textbf{0.826} 	&\textbf{0.821} 	&\textbf{0.032} 
&\textbf{0.737} 	&\textbf{0.661} 	&\textbf{0.076}  \\

				\bottomrule
			\end{tabular}}

	\end{center}
	\label{ab-psem}
\end{table}

\begin{itemize}
    \item \textbf{w/o Sem} indicates that semantic embedding is removed from the PSE module, leaving only position embedding.
    \item \textbf{w/o Pos} indicates that positional embedding is removed from the PSE module, leaving only learnable semantic embedding. 
    \item \textbf{tra-fix} represents the replacement of fixation generated by TASED-Net \cite{DBLP:conf/iccv/MinC19} with fixation extracted using a traditional method GBVS \cite{DBLP:conf/nips/HarelKP06}.
\end{itemize}

\begin{figure}[!t]
    \centering
    \includegraphics[scale=0.36]{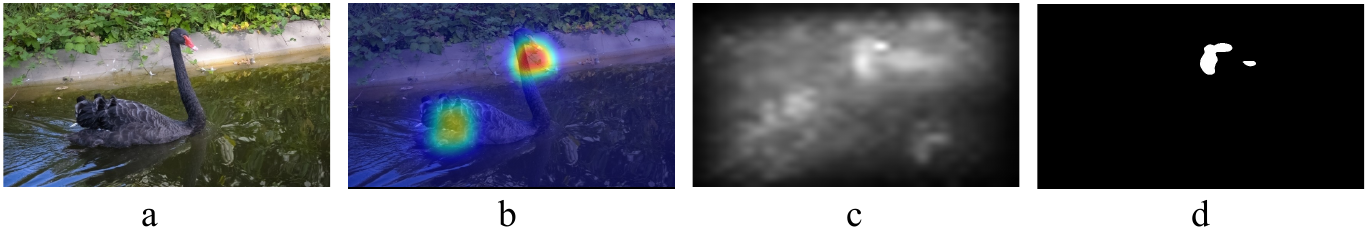}
    \caption{a: Image. b: Fixations generated by the TASED-Net (deep learning based method). c: Fixations generated by the GBVS (traditional method) d: Fixations filtered by a threshold using GBVS.}
    \label{fig:gbvs}
\end{figure}

First, to verify the the effectiveness of the PSE module, we design experiments that focus on its two main components: semantic information and positional information. Positional information is obtained from fixation annotations, which provides basic localization information to the network. On the other hand, semantic information is comprised of randomly initialized semantic embedding that gradually learn target-specific semantics during training. These two components are jointly used to initialize the object query. Experiment \textbf{w/o Sem} from Table \ref{ab-psem} highlights the importance of semantic information, showing that with the help of semantic embedding on the DAVIS, the MAE score reaches 8.5\%. The positional embedding, which provides the main information from the fixation, significantly improves the performance of the network. As shown in the experiment \textbf{w/o Pos}, the positional embedding improves S-measure by 4.03\%, F-measure by 4.85\%, and MAE score by 33.3\% on the DAVIS dataset. It is worth noting that, given the random initialization of semantic embedding, the experiment \textbf{w/o Pos} essentially eliminates the performance associated with fixation information.

Second, because we use a deep-learning based method to predict fixation as input for training our EGCNet for VSOD tasks, it is possible that this method may introduce additional implicit supervised information to the network. This is because the deep learning method relies on supervised training using ground-truth fixation data. To demonstrate that the effectiveness of our overall approach does not depend solely on the deep learning method, we designed Experiment \textbf{tra-fix}, which retrains the network using fixations predicted by the traditional method GBVS method. The predicted results of the GBVS method are shown in Fig. \ref{fig:gbvs}(c). To obtain more accurate results, we applied threshold filtering to the GBVS predictions, as shown in Fig. \ref{fig:gbvs}(d). As shown in Table \ref{ab-psem}, our proposed method, which is based on GBVS predictions, still outperform the second-best method in Table \ref{compare_table}. The performance of our model is 3.22\% and 11.7\% better than the second-best model (\ie, WVSODP) in terms of F-measure and MAE score on the DAVSOD dataset.

\begin{table}[!t]
	\scriptsize
	\caption{Quantitative ablation evaluation of sampling method for fixation annotations on the VSOD datasets DAVIS and DAVSOD. Black bold fonts indicate the best performance.}
	\begin{center}
				 \renewcommand\arraystretch{1.5}
		\setlength{\tabcolsep}{1.3mm}{
		\vspace{0.2cm}
			\begin{tabular}{c|ccc|ccc}
			\toprule
\multicolumn{1}{c|}{\multirow{2}{*}{Method}} & \multicolumn{3}{c|}{\textbf{DAVIS}}  & \multicolumn{3}{c}{\textbf{DAVSOD}}                                                                                                                                                                 \\ \cline{2-7} 

\multicolumn{1}{c|}{}
&\multicolumn{1}{c}{$S_{\alpha}\uparrow$} 
&\multicolumn{1}{c}{$F_{\beta}\uparrow$} & \multicolumn{1}{c}{${MAE}\downarrow$} 
& \multicolumn{1}{c}{$S_{\alpha}\uparrow$}  
&\multicolumn{1}{c}{$F_{\beta}\uparrow$}
& \multicolumn{1}{c}{$MAE\downarrow$} \\ \toprule
{Argmax}
& {0.811} & {0.812} & {0.037}
&  & {0.651} & {0.083} \\
Box
& 0.816  & 0.803	& 0.034
& 0.731  & 0.653    & 0.079      \\
Random point
& 0.813  & 0.797	& 0.036
& 0.727  & 0.646    & 0.083     \\
Center point
&\textbf{0.826} 	&\textbf{0.821} 	&\textbf{0.032}   
&\textbf{0.737} 	&\textbf{0.661} 	&\textbf{0.076}  \\

				\bottomrule
			\end{tabular}}

	\end{center}
	\label{psem-sample}
\end{table}

Finally, we validate the sampling method for fixation annotation in the PSE module. As the fixations in each frame of a video vary in size, it is challenging to sample the same number of points for all frames, making the process time-consuming. To address this challenge, we use the geometric centers of fixations as the sampling point for position encoding. {We conduct experiments with three additional sampling methods: using the minimum bounding box of fixations for positional encoding, randomly sampling fixations, and selecting the point with the highest probability value. As demonstrated in Table \ref{psem-sample}, our \textbf{center point} sampling method achieves the best performance. For example, our \textbf{center point} sampling method achieves better performance in terms of MAE score on the DAVSOD dataset than the \textbf{box} method by 3.8\% and the \textbf{random point} method by 8.4\%. In contrast, \textbf{argmax} performs the least well on the DAVIS dataset, where its MAE increases from 0.032 to 0.037.}

\subsubsection{Effectiveness of SLQ Competitor}
We conduct experiments to verify the effectiveness of our SLQ Competitor, as show in Table \ref{ab-sql}. The specific experimental settings are as follows:
\begin{itemize}
    \item \textbf{w/o $Score_{sem}$} represents the removal of object query selection scores $Score_{sem}$ for semantic feature similarity. 
    \item \textbf{w/o $Score_{loc}$} represents the removal of object query selection scores $Score_{loc}$ for target localization matching.
    \item \textbf{+Hungarian} means that we replace our proposed method with the original bipartite graph algorithm to compare it with the Hungarian algorithm used in the original network.
\end{itemize}

\begin{figure*}[!t]
    \centering
    \includegraphics[scale=0.45]{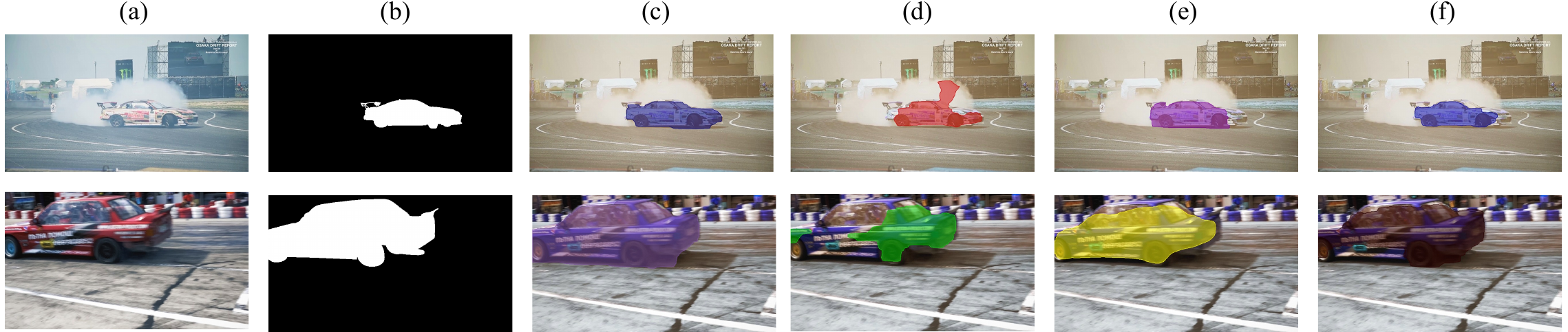}
    \caption{Visual comparison of ablation study for SLQ Competitor. (a) Image. (b) GT. (c) FULL. (d) w/o $Score_{sem}$. (e) w/o $Score_{loc}$. (f) +Hungarian}
    \label{fig:ab_sql_fig}
\end{figure*}

In SLQ competitor, we have built semantic score $Score_{sem}$ and locality score $Score_{loc}$ to evaluate the accuracy and completeness of target detection in object queries. These scores are based on the two main pieces of information contained in object query, semantic information, and target location information. To verify the effectiveness of our object query selection measurements, we conduct ablative experiments on these two parts respectively. 
The semantic score represents the main evaluation score for object queries, and it has a significant impact on the network's prediction results due to the use of sparse labels (scribble labels). By utilizing the semantic score, we can identify more complete and accurate predicted object query. Experiment \textbf{w/o $Score_{sem}$} in Table \ref{ab-sql} demonstrates the effectiveness of the semantic feature similarity score. 
The S-measure increases from 0.719 to 0.737, F-measure increases from 0.632 to 0.661, and the MAE score increases from 0.094 to 0.076 on the DAVSOD dataset when compared to SLQ Competitor without $Score_{sem}$, representing a growth of 2.5\%, 4.6\%, and 19.1\% respectively.

\begin{table}[!t]
	\scriptsize
	\caption{Quantitative ablation evaluation of SLQ Competitor on the VSOD datasets DAVIS and DAVSOD. Black bold fonts indicate the best performance.}
	\begin{center}
				 \renewcommand\arraystretch{1.5}
		\setlength{\tabcolsep}{1.25mm}{
		\vspace{0.2cm}
			\begin{tabular}{c|ccc|ccc}
			\toprule
\multicolumn{1}{c|}{\multirow{2}{*}{Method}} & \multicolumn{3}{c|}{\textbf{DAVIS}}  & \multicolumn{3}{c}{\textbf{DAVSOD}}                                                                                                                                                                 \\ \cline{2-7} 

\multicolumn{1}{c|}{}
&\multicolumn{1}{c}{$S_{\alpha}\uparrow$} 
&\multicolumn{1}{c}{$F_{\beta}\uparrow$} & \multicolumn{1}{c}{${MAE}\downarrow$} 
& \multicolumn{1}{c}{$S_{\alpha}\uparrow$}  
&\multicolumn{1}{c}{$F_{\beta}\uparrow$}
& \multicolumn{1}{c}{$MAE\downarrow$} \\ \toprule
w/o $Score_{sem}$
& 0.796	     & 0.783	 & 0.042
&0.719	     & 0.632	 & 0.094 \\
w/o $Score_{loc}$
& 0.810      & 0.801	 & 0.034
& 0.725      & 0.643	 & 0.085 \\
+Hungarian
& 0.813	     & 0.803	& 0.035
& 0.723	     & 0.647	& 0.088 \\
FULL
&\textbf{0.826} 	&\textbf{0.821} 	&\textbf{0.032}  
&\textbf{0.737} 	&\textbf{0.661} 	&\textbf{0.076}  \\

				\bottomrule
			\end{tabular}}

	\end{center}
	\label{ab-sql}
\end{table}

In addition, the locality score $Score_{loc}$ also plays a crucial role in object query selection. Relying solely on similarity to match object query is not always reliable as it might lead to misclassification, especially when foreground and background object features are similar, resulting in a high background score for the object query. In such cases, locality score becomes crucial.
To validate the effectiveness of the locality score, we conduct a experiment \textbf{w/o $Score_{loc}$}. The results in the Table \ref{ab-sql} reveal that with the use of the locality score, the F-measure, S-measure, and MAE score improve by 1.7\%, 2.8\%, and 10.6\%, respectively, on the DAVSOD dataset. This highlights the significance of the locality score in accurate object query selection.

The Deformable DETR model leverages bipartite matching to solve for multi-to-multi problems, where multiple object queries correspond to multiple ground-truths. However, our task is a multi-to-one task, and to address it more effectively, we propose SLQ competitor. To evaluate the effectiveness of our method compared to the original matching method, we replace our matching algorithm with the Hungarian matching algorithm and conduct a \textbf{+Hungarian} experiment.
The results in Table \ref{ab-sql} demonstrate that our method achieve better performance in the VSOD task compared to the Hungarian matching algorithm. Specifically, on the DAVSOD dataset, our matching algorithm perform better than Hungarian matching in S-measure, F-measure, and MAE score by 1.9\%, 2.2\%, and 13.6\%, respectively.

In addition, we also provide some visualization results in Fig. \ref{fig:ab_sql_fig} to demonstrate the role of the SLQ Competitor. From the figure, we can observe that without the support of semantic score and locality score, the integrity and localization accuracy of the network's predictions are affected. Moreover, compared to traditional binary matching algorithms, our algorithm performs better in scenes with lower contrast, providing more complete and accurate predictions.

\subsubsection{Effectiveness of IIMC Model}
\textbf{Contrastive Stage.} Firstly, we verify the overall contrastive satge of our IIMC module, which is composed of two stages: Intra-video Contrast stage and Inter-video Contrast stage. We set up two experiments in Table \ref{ab-iimc} to demonstrate the individual contributions of these two stages with the following experimental settings:

\begin{itemize}
    \item \textbf{w/o intra Sta.} means that we remove the intra-video contrast stage, and rely solely on the features between different videos for contrastive training.
    \item \textbf{w/o inter Sta.} means that we remove the inter-video comparison stage and only use the foreground and background features between key frame and reference frame of the same video for contrasting during training.
\end{itemize}

\begin{figure}[!t]
    \centering

    \includegraphics[scale=0.27]{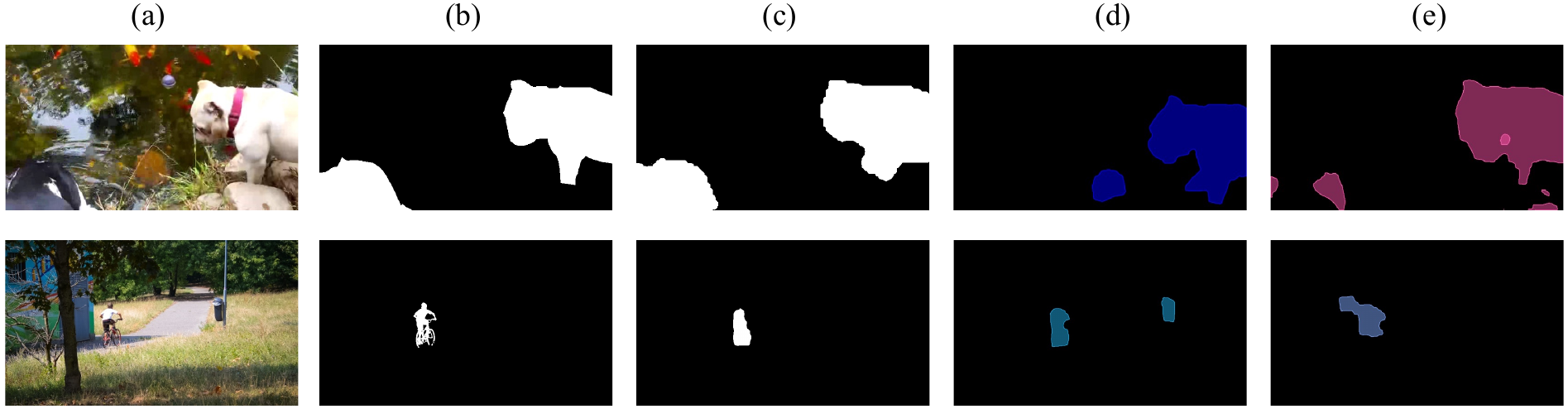}
    \caption{Visual comparison of ablation study for IIMC model. (a) Image. (b) GT. (c) FULL. (d) w/o intra Sta. (e) w/o inter Sta.}
    \label{fig:ab_iimc_fig}
\end{figure}

\begin{table}[!t]
	\scriptsize
	\caption{Quantitative ablation evaluation of IIMC model on the DAVIS and DAVSOD datasets. Black bold fonts indicate the best performance.}
 \vspace{-0.5cm}
	\begin{center}
				 \renewcommand\arraystretch{1.5}
		\setlength{\tabcolsep}{1.4mm}{
		\vspace{0.2cm}
			\begin{tabular}{c|ccc|ccc}
			\toprule
\multicolumn{1}{c|}{\multirow{2}{*}{Method}} & \multicolumn{3}{c|}{\textbf{DAVIS}}  & \multicolumn{3}{c}{\textbf{DAVSOD}}                                                                                                                                                                 \\ \cline{2-7} 

\multicolumn{1}{c|}{}
&\multicolumn{1}{c}{$S_{\alpha}\uparrow$} 
&\multicolumn{1}{c}{$F_{\beta}\uparrow$} & \multicolumn{1}{c}{${MAE}\downarrow$} 
& \multicolumn{1}{c}{$S_{\alpha}\uparrow$}  
&\multicolumn{1}{c}{$F_{\beta}\uparrow$}
& \multicolumn{1}{c}{$MAE\downarrow$} \\ \toprule
w/o intra Sta.
& 0.775	     & 0.769	 & 0.050	
& 0.694	     & 0.617	 & 0.099\\
w/o inter Sta.
& 0.784	     & 0.774	 & 0.045
& 0.703	     & 0.625	 & 0.090\\
FULL
&\textbf{0.826} 	&\textbf{0.821} 	&\textbf{0.032}  
&\textbf{0.737} 	&\textbf{0.661} 	&\textbf{0.076}  \\

				\bottomrule
			\end{tabular}}

	\end{center}
	\label{ab-iimc}
\end{table}

We propose the IIMC module, which utilizes contrastive learning to learn temporal information from videos by comparing the similarity of foreground features and different foreground-background features across frames. This module contains two key contrastive stages: intra-video contrast stage and inter-video contrast stage. Firstly, the intra-video contrast stage provides the network with more discriminative pixel-level features, learns more detailed local features, and enables the module to obtain more complete salient objects. To verify the effectiveness of the intra-video contrast stage, we design an experiment \textbf{w/o intra Sta.} Results listed in Table \ref{ab-iimc} demonstrate that with the aid of the intra-video contrast stage, our IIMC module achieves remarkable performance improvements. For example, on the DAVIS dataset, the F-measure improves from 0.769 to 0.821, with a 6.8\% increase. The S-measure improves from 0.775 to 0.826, with a 6.6\% increase, and the MAE score drops by 0.018, which is a 36\% improvement compared to the experiment without intra-video contrast stage.

Secondly, the inter-video contrast stage in the video provides more precise information about salient objects for the network by comparing the features of foreground objects across different videos. We also conduct experiments \textbf{w/o inter Sta.} to demonstrate the significance of this stage. As shown in Table \ref{ab-iimc}, our inter-video contrast stage results in considerable improvements for the network. For example, on the DAVSOD dataset, our S-measure, F-measure, and MAE score increase by 4.8\%, 5.8\%, and 15.6\%, respectively, compared to the network without the inter-video contrast stage.

Additionally, in the visualization shown in Fig. \ref{fig:ab_iimc_fig}, we can find that, with the support of the intra-video contrast stage, we have utilized the sequential relationship between key frame and reference frame to exclude false positive salient objects (such as a trash can). We can also observe in the figure that through the inter-video contrast stage, we provide more negative examples for contrastive learning, thereby completing more pixels of the salient objects.\\
{
\textbf{Memory Bank.}
Secondly, we verify the important component of the IIMC module, the Memory Bank. We set up two experiments in Table \ref{ab-memory_bank} to separately verify the frame-level Memory Bank and the video-level Memory Bank, using the following settings:
\begin{itemize}
    \item \textbf{w/o Frame MB} means that we remove the Frame-level Memory Bank in the intra-video contrast stage to mainly verify the importance of inter-frame spatiotemporal features.
    \item \textbf{w/o Vdieo MB}  means that we remove the Video-level Memory Bank in the inter-video contrast stage to mainly verify the importance of cross-video contrast features.
\end{itemize}

\begin{table}[!t]
    \scriptsize
    \caption{Quantitative ablation evaluation of Memory Bank on the DAVIS and DAVSOD datasets. Black bold fonts indicate the best performance.}

    \begin{center}
        \renewcommand\arraystretch{1.5}
        \setlength{\tabcolsep}{1.4mm}
        \vspace{0.2cm}
        \begin{tabular}{c|ccc|ccc}
            \toprule
            \multicolumn{1}{c|}{\multirow{2}{*}{Method}} &
            \multicolumn{3}{c|}{\textbf{DAVIS}}  &
            \multicolumn{3}{c}{\textbf{DAVSOD}} \\ \cline{2-7}
            & $S_{\alpha}\uparrow$ & $F_{\beta}\uparrow$ & ${MAE}\downarrow$ 
            & $S_{\alpha}\uparrow$ & $F_{\beta}\uparrow$ & $MAE\downarrow$ \\
            \toprule
            w/o Frame MB & 0.787 & 0.792 & 0.037 & 0.709 & 0.638 & 0.086 \\
            w/o Video MB & 0.809 & 0.812 & 0.035 & 0.719 & 0.649 & 0.080 \\
            FULL & \textbf{0.826} & \textbf{0.821} & \textbf{0.032} 
                 & \textbf{0.737} & \textbf{0.661} & \textbf{0.076} \\
            \bottomrule
        \end{tabular}
    \end{center}
    \label{ab-memory_bank}
\end{table}

To enhance the feature representation capability of contrastive learning, it's typically necessary to introduce more negative samples for comparison. We also use a memory bank to store the feature representations of the current video and other videos as negative examples for feature contrast learning. First, for temporal feature learning between videos, we store the background feature vector of the current video as a negative sample relative to the foreground feature vector, to learn the spatiotemporal information between frames. Experiment \textbf{w/o Frame MB} demonstrates the importance of the Frame-level Memory Bank for the network. For example, on the DAVIS dataset, with the aid of the Frame-level Memory Bank, the S-measure, F-measure, and MAE improve by 5.0\%, 3.7\%, and 13.5\%, respectively.
Secondly, to learn more distinctive features and increase scene diversity, we set up a Video-level Memory Bank to store representative foreground features of multiple videos. For the current video, the feature vectors of other videos stored in the memory bank will serve as negative samples for contrastive learning. Experiment \textbf{w/o Vdieo MB} confirms the importance of our Video-level Memory Bank. For example, on the DAVSOD dataset, compared to the version without Video-level Memory Bank, we improve the S-measure, F-measure, and MAE score by 2.5\%, 1.8\%, and 5.0\%, respectively.
}

{
\subsubsection{Analysis of Sampling Interval}
In the field of VSOD, most methods adopt consecutive multi-frame input rather than interval input. Regarding the issue of sampling interval, we didn't sample adjacent frames as keyframes and reference frames, but randomly sampled within 0 to 10 frames nearby. To this end, we set up two experiments to compare the sampling of larger ranges (within 20 frames) and smaller ranges (within 5 frames). As shown in Table \ref{ab-sample}, For the 5-frame range, while this setting minimizes the risk of losing or occluding the salient object, the frames tend to be overly similar, leading to reduced contrast between foreground and background. As a result, the effectiveness of contrastive learning is limited. In contrast, the 20-frame range introduces more pronounced changes, but the salient object’s position or appearance may shift significantly or become partially occluded. These large variations introduce noisy supervision signals, slightly degrading performance compared to our final configuration.

\begin{table}[!t]
    \scriptsize
    \caption{Quantitative ablation evaluation of video sampling intervals on the VSOD datasets DAVIS and DAVSOD. Black bold fonts indicate the best performance.}
    \begin{center}
        \renewcommand\arraystretch{1.5}
        \setlength{\tabcolsep}{2.0mm}{
        \vspace{0.2cm}
        \begin{tabular}{c|ccc|ccc}
            \toprule
            \multicolumn{1}{c|}{\multirow{2}{*}{Method}} &
            \multicolumn{3}{c|}{\textbf{DAVIS}} &
            \multicolumn{3}{c}{\textbf{DAVSOD}} \\ \cline{2-7}
            &
            $S_{\alpha}\uparrow$ & $F_{\beta}\uparrow$ & ${MAE}\downarrow$ &
            $S_{\alpha}\uparrow$ & $F_{\beta}\uparrow$ & $MAE\downarrow$ \\
            \toprule
            5 frame &
            0.819 & 0.818 & 0.035 &
            0.728 & 0.660 & 0.078 \\
            20 frame &
            0.821 & 0.810 & 0.033 &
            0.731 & 0.657 & 0.084 \\
            10 frame (ours) &
            \textbf{0.826} & \textbf{0.821} & \textbf{0.032} &
            \textbf{0.737} & \textbf{0.661} & \textbf{0.076} \\
            \bottomrule
        \end{tabular}}
    \end{center}
    \label{ab-sample}
\end{table}

}




\section{Conclusion}
We approach VSOD task from a fresh perspective, introducing a related field (VSP task) to assist us in better results. To this end, we propose the EGCNet. We start by exploring how to utilize fixations and then a PSE moudle, incorporating semantic and location constraints, is introduced to handle the priors brought by these new information. In addition, under the current task setting, we perform spatiotemporal feature modeling through two consecutive steps: feature selection and feature contrast.
This allows us to achieve precise salient object detection even under sparse supervision. Our proposed EGCNet outperforms other 5 state-of-the-art weakly supervised methods and even rivals fully supervised methods on certain datasets. 
It also provides a solid foundation for various downstream tasks.

 
%

\bibliography{ref}
\bibliographystyle{IEEEtran}

\vfill

\end{document}